\pdfoutput=1
\documentclass[11pt]{article}
\usepackage[]{acl} 
\usepackage{times}
\usepackage{latexsym}
\usepackage[T1]{fontenc}
\usepackage[utf8]{inputenc}
\usepackage{microtype}
\usepackage{inconsolata}

\usepackage{graphicx}
\usepackage{adjustbox}
\usepackage{booktabs}
\usepackage{fixltx2e}
\usepackage{enumitem}
\usepackage{amssymb}
\usepackage{amsmath}
\usepackage[most]{tcolorbox}
\usepackage{bbm}
\usepackage{xspace}
\usepackage{colortbl}
\usepackage{titlesec}

\newcommand{\our}{\textsc{MetaEntail-RE}\xspace}

\newcommand{\rb}{\rotatebox}
\newcommand{\bb}{\textbf}
\newcommand{\uu}{\underline}
\newcommand{\cc}{\cellcolor}
\newcommand{\sst}{\textsubscript}
\definecolor{block-gray}{gray}{0.85}
\newtcolorbox{blockquotec}{colback=block-gray,grow to right by=-1mm,grow to left by=-1mm,boxrule=0pt,boxsep=0pt,breakable}

\newcommand\y{0} 

\title{Entangled Relations: Leveraging NLI and Meta-analysis\\to Enhance Biomedical Relation Extraction}
\author{ 
    William Hogan $\qquad$ Jingbo Shang\thanks{$\,$ Corresponding author} \\
    Department of Computer Science \& Engineering\\
    University of California, San Diego \\
    \texttt{\{whogan,jshang\}@ucsd.edu}}
\begin{document}
\maketitle

\begin{abstract}
Recent research efforts have explored the potential of leveraging natural language inference (NLI) techniques to enhance relation extraction (RE). 
In this vein, we introduce \our, a novel adaptation method that harnesses NLI principles to enhance RE performance. 
Our approach follows past works by verbalizing relation classes into class-indicative hypotheses, aligning a traditionally multi-class classification task to one of textual entailment. We introduce three key enhancements: 
(1) Meta-class analysis which, instead of labeling non-entailed premise-hypothesis pairs with the less informative ``neutral'' entailment label, provides additional context by analyzing overarching meta-relationships between classes; 
(2) Feasible hypothesis filtering, which removes unlikely hypotheses from consideration based on domain knowledge derived from data; and 
(3) Group-based prediction selection, which further improves performance by selecting highly confident predictions. 
\our is conceptually simple and empirically powerful, yielding significant improvements over conventional relation extraction techniques and other NLI formulations. We observe surprisingly large F1 gains of 17.6 points on BioRED and 13.4 points on ReTACRED compared to conventional methods, underscoring the versatility of \our across both biomedical and general domains.
\end{abstract}

\section{Introduction}\label{sec:introduction}

Relation extraction (RE) is an NLP task that distills factual information from text by identifying relationships between entities in the form of fact triplets (e.g.,⟨head, relation, tail⟩) \cite{Califf1997RelationalLO, Mintz2009DistantSF, BaldiniSoares2019MatchingTB, Wan2023GPTREIL}. RE facilitates various downstream applications such as knowledge graph construction, question answering, and information retrieval ~\cite{Yuan2022JointME, He2023EntityRA, Yamada2023BiomedicalRE}; however, creating datasets for training RE models is costly and challenging, requiring annotators to identify entities and relations across large sections of text \cite{docred, biored}.

Recent efforts have explored adapting the RE task into a natural language inference (NLI) task, where the goal is to determine whether a given hypothesis logically follows from, contradicts, or is neutral with respect to a premise. This adaptation enables the use of relatively large NLI datasets to improve performance on an RE-adapted task \cite{Sainz2021LabelVA, Sainz2022TextualEF, nbr}. RE-to-NLI works transform relation instances into premises paired with $m$ class-indicative hypotheses where $m$ is the number of relation classes in a dataset. A language model is trained to label premise-hypothesis pairs as \textit{entailed}, \textit{contradicted}, or \textit{neutral}. We build on this work by introducing \our, a novel NLI adaptation method that improves RE performance by leveraging three key enhancements: automatic feasible hypothesis filtering, meta-class analysis, and group-based prediction selection.

\textbf{Feasible hypothesis filter}: We first introduce a feasible hypothesis filter that automatically removes infeasible hypotheses based on domain knowledge derived from data. To develop this filter automatically, we approximate valid sets of entity-type pairs corresponding to each relation class by aggregating all relations in the training data. These approximated sets of valid type-pairs are then used to remove hypotheses that verbalize infeasible relationships. For instance, in the BioRED dataset \cite{biored}, it is impossible for a \textit{gene} to ``bind'' to a \textit{disease} (i.e., the ``bind'' label is not applicable to gene-disease entity-type pairs). We therefore remove the ``bind'' hypothesis from all instances with gene-disease entity types.  This filter improves training efficiency by reducing the number of NLI instances.

\textbf{Meta-class analysis}: In past RE-to-NLI works, if a premise does not entail a hypothesis, the corresponding NLI label assigned is ``neutral'' \cite{Sainz2021LabelVA, nbr}. However, this misses an implicit training signal we can gain by analyzing the semantics of a dataset's relation classes. When assigning NLI labels to adapted RE instances, we distinguish between \textit{task-based} mutual exclusivity and \textit{definition-based} mutual exclusivity. Task-based mutual exclusivity is an artifact of the single-class classification task inherent to a dataset. Each input instance is annotated with a single relation class, thereby arbitrarily making all classes mutually exclusive. In contrast, definition-based mutual exclusivity is derived from definitions of relation classes. For example, within the BioRED dataset \cite{biored}, the ``positive correlation'' class is definitionally mutually exclusive and contradictory to the ``negative correlation'' class \cite{biored}. 

If two classes are definitionally mutually exclusive, we apply the ``contradict'' label to the appropriate premise-hypothesis pair, thereby injecting additional information about the meta-relationship between relation classes which the model can exploit while learning relationship representations. Leveraging this insight, we can glean multiple informative training signals from a single relation instance when adapting the relation extraction task into the natural language inference task. We call this method meta-class analysis (MCA) and use it to determine the appropriate NLI labels for each premise-hypothesis pair. We show through ablation experiments that MCA leads to significant gains on an RE-adapted task.

\textbf{Group-based prediction selection}: Group-based prediction selection exploits the feature of RE-to-NLI adaptation in that each relation instance is converted into a group of premise-hypothesis pairs where each hypothesis verbalizes a relation class in the dataset. When evaluating cases where the model predicts multiple ``entail'' labels within a single group, we can select the most confident ``entail'' prediction and ignore other predictions. Our results demonstrate that this group-based prediction selection method leads to additional gains.

\our as an RE-to-NLI adaptation method is technically domain agnostic; however, it is particularly well-suited for biomedical RE where associations often have opposing classes such as ``positively correlated'' and ``negatively correlated'' \cite{biored} or ``agonist'' and ``antagonist'' \cite{Taboureau2010ChemProtAD} enabling a rich MCA. We also find that associations in biomedical RE are often type-dependent compared to general domain RE, making the feasible hypothesis filter more effective at trimming infeasible hypotheses. Still, we extend our evaluations beyond the biomedical domain to determine how \our fares on general domain RE datasets. Notably, we observe improvements in both domains, reinforcing the effectiveness and versatility of \our. We summarize the main contributions of this work as the following:
\begin{itemize}[nosep,leftmargin=*]    
    \item We introduce a novel RE-to-NLI adaptation method, \our, and showcase its robustness and versatility in RE datasets from general and biomedical domains.
    \item We illustrate through ablation experiments the effectiveness of components of \our. 
    \item We openly provide all code, experimental settings, and datasets used to substantiate the claims made in this paper.\footnote{\url{https://github.com/wphogan/metaentail-re}}
\end{itemize}

\begin{figure*}[t]
    \centering
    \includegraphics[width=1.0\textwidth]{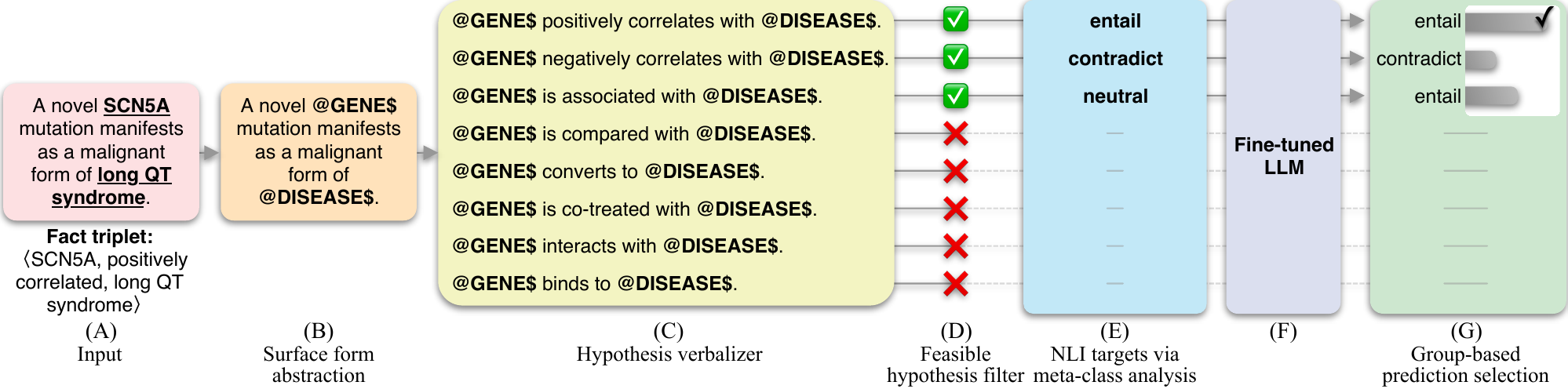}
    \caption{
    Data flow used for \our. The original RE input instance (A) is converted into a premise where surface forms are masked with corresponding entity types (B). Each relation class is verbalized into a hypothesis (C), and a feasible hypothesis filter (D) removes infeasible hypotheses based on the pair of entity types. NLI labels are generated via meta-class analysis (E), which are the labels used to fine-tune an LLM via cross-entropy (F). Finally, we use softmax probabilities as a proxy for the model's confidence and select the most confident ``entail'' prediction among the group of predictions (G). Note that the model makes three predictions in this example—one for each feasible hypothesis. The second ``entail'' prediction is incorrect but the group-based prediction module selects the first and correct ``entail'' prediction by assessing the model's confidence.
    } \label{fig:1}
\end{figure*}

\section{Related Work}\label{sec:related_work}
Traditionally, RE has been approached as a classification task, where input instances are classified as belonging to a relational class \cite{Califf1997RelationalLO, Mintz2009DistantSF,BaldiniSoares2019MatchingTB, Wan2023GPTREIL}. These methods have several drawbacks: they tend to generalize poorly \cite{Peng2020LearningFC, nbr}, and they heavily rely on relatively small and disjoint  RE datasets. To account for these drawbacks, recent works have proposed clever adaptation methods to recast RE into adjacent NLP tasks, such as a question-answering \cite{levy-etal-2017-zero} and NLI \cite{obamuyide-vlachos-2018-zero, Sainz2021LabelVA, Sainz2022TextualEF, nbr}. Task adaptation presents an opportunity to leverage the relatively large datasets available for other tasks (e.g., SQuAD \cite{rajpurkar-etal-2016-squad}, MultiNLI \cite{mnli}, SNLI \cite{bowman-etal-2015-large}, etc.), which can be particularly advantageous in the context of biomedical RE where datasets are often limited. 

\citet{levy-etal-2017-zero} recast RE into a question-answering task by associating relation instances with one or more natural-language questions, resulting in predicted spans denoting class indicative text. \citet{obamuyide-vlachos-2018-zero} adapts general domain RE into an NLI task by using relation instances as premises where each premise is paired with a hypothesis generated by verbalizing a relation class. In doing so, they formulate a binary entailment task where they predict whether or not a premise entails the corresponding hypothesis.

\citet{Sainz2021LabelVA} expands on \citet{obamuyide-vlachos-2018-zero} by incorporating a three-label classification objective where a model can predict \textit{entail}, \textit{contradict}, and \textit{neutral} depending on the premise-hypothesis pair, bringing the task in line with a standard NLI formulation \cite{Dagan2005ThePR}. They manually generate hypothesis templates corresponding to each relation class in a dataset, and NLI labels are assigned based on the alignment of the premise-hypothesis pair. If the corresponding hypothesis is the verbalized version of the ground truth relation label, then ``entail'' is assigned as the NLI label for the instance. The ``neutral'' label is applied to positive class hypotheses which do not align with a given premise. The ``contradict'' label is applied in two cases: (1) if the premise is a positive relation instance (e.g., any class other than ``no relation''), the ``no-relation'' hypothesis is labeled as ``contradict,'' and (2) if the premise is a negative instance (e.g., ``no relation''), then all other positive class hypotheses are labeled as ``contradict.'' \citet{Sainz2021LabelVA} fine-tune a language model pre-trained on the MultiNLI \cite{mnli} dataset to predict generated NLI labels. They observe impressive results in zero- and few-shot scenarios on TACRED \cite{zhang2017tacred}, a general domain, sentence-level RE dataset.

\citet{nbr} explores cross-domain transfer learning, leveraging indirect supervision from general domain NLI datasets to improve biomedical RE-to-NLI adapted methods. Our work can be considered an extension of their proposed NBR method. However, we introduce a few key improvements: meta-class analysis, a feasible hypothesis filter, and group-based prediction selection. We also expand evaluations beyond sentence-level RE to include more challenging document-level RE \cite{Li2016BioCreativeVC, biored}.
\section{Problem Statement}\label{sec:problem_statement}
Our problem is a hybridization of RE and NLI; as such, we describe both tasks, as well as the adapted RE-to-NLI task.

\textbf{Relation Extraction (RE)}: RE takes inputs $\{x_1, x_2,\ldots, x_n\} \in X_{\textsc{RE}}$ where $X_{\textsc{RE}}$ is a corpus of sentences, paragraphs, or documents of size $n$ and $x_i$ is a singular instance containing an entity pair $e_{i_{1}}$ and $e_{i_{2}}$. Each input $x_i$ has a corresponding label $y_i$. Labels $\{y_1, y_2,\ldots, y_n\} = Y_{\textsc{RE}}$ belong to a set of $m$ relation classes $R = \{r_1, r_2,\ldots, r_m\}$. RE seeks to identify which class links the co-mentioned entities to form a fact triplet $\langle e_{i_{1}}, y_i, e_{i_{2}} \rangle$, or, semantically, ⟨\textit{head}, \textit{relation}, \textit{tail}⟩.

\textbf{Natural Language Inference (NLI)}: NLI takes a premise $p_i \in P$ and a hypothesis $h_i \in H$, where $P$ and $H$ are the set of premises and hypotheses in a corpus, respectively, and seeks to determine whether the premise entails, contradicts, or is neutral to the respective hypothesis \cite{Dagan2005ThePR, bowman-etal-2015-large}. 
Using $\hat{y}_i$ to represent an NLI label applied to the $i_{th}$ instance, $\hat{y}_i \in \{entail, contradict, neutral\}$, and a single NLI example can be expressed as $\langle p_i, \hat{y}_i, h_i \rangle$. 

\textbf{RE-to-NLI Adaptation}: RE-to-NLI adaptation converts RE inputs and labels into premise-hypothesis pairs such that each input instance maps to $|R|$ premise-hypotheses pairs: $(x_i,y_i)\rightarrow\{(p_i, \hat{y}_j, h_j)\}_{j=1}^{|R|}$. We decompose RE-to-NLI adaptation into the following sub-steps:

\begin{enumerate}[label=(\alph*),nosep,leftmargin=*]    
\item Premise generation, $x_i\rightarrow p_i$: Input instances $x_i \in X_{\textsc{RE}}$ directly become premises $p_i \in P^{|X_{\textsc{RE}}|}$ where $P$ is the collection of all premises generated from $X_{\textsc{RE}}$.
\item Hypothesis generation, $H_{i} = \{h_j\}_{j=1}^{|R|}$: In the hypothesis generation step, a set of hypotheses $H_{i}$ paired with each premise $p_i$. This is achieved by first verbalizing relation classes in $R$ into a set of $m$ hypothesis templates $T=\{t_1, t_2,\ldots, t_m\}$. Each hypothesis template contains head and tail entity placeholders, which are replaced by the head and tail entities found in the corresponding premise $p_i$. The verbalizer function $f_{verbalizer}(\cdot)$ takes each hypothesis template and entity pair in premise $p_i$ to produce the set of hypotheses $H_i = \{f_{verbalizer}(t_j, e_{i_{1}}, e_{i_{2}})\}_{j=1}^{|R|}$. 
\item NLI label generation, $\hat{Y} = \{\hat{y}_i\}_{i=1}^{|X_\textsc{RE}| \times |R|}$: The set of NLI labels $\hat{Y}$ is generated via a function which takes the original instance label $y_i$ and the premise-hypothesis pair $f_{target}(y_i, p_i, h_j) \rightarrow \hat{y}_j$ where NLI label $\hat{y}_j = entail$ iff verbalized class-indicative hypothesis $h_j$ aligns with the ground truth label $y_i$, and, depending on the adaptation method, $\hat{y}_j$ is assigned $neutral$ or $contradict$ for non-aligned hypotheses. 
\end{enumerate}

The RE-to-NLI task is to correctly predict entailed premise-hypothesis pairs where each entailed pair has a 1-to-1 mapping to the original RE label.

%
%







\section{Methods}\label{sec:methods}
This section sequentially discusses the modules used in \our (see Figure \ref{fig:1}).

\textbf{Premise Construction:} Following \citet{nbr}, a relation instance $x_i$ is transformed into a premise by replacing surface forms of the subject and object entities, $e_1$ and $e_2$, respectively, with their corresponding entity types, $e_{1_{type}}$ and $e_{2_{type}}$. Abstracting entity surface forms into entity types helps alleviate the long-tail nature of biomedical entities and encourages language models to learn from context instead of shallow heuristics \cite{Peng2020LearningFC}. The start and end spans of entity types are denoted with ``$@$'' and ``$\$$,'' respectively. 

\textbf{Hypothesis Verbalizer:} Past works have manually generated hypothesis templates for each relation class in a dataset which are then used, in turn, to generate hypotheses to pair with a given premise. A secondary contribution of \our is that we reduce this human effort by leveraging LLMs to automatically generate the set of hypothesis templates $\{t_1, t_2,\ldots,t_m\}\in T$, where $m$ corresponds to the number of relation classes in a dataset. We prompt an LLM\footnote{We use ChatGPT (GPT 3.5) via OpenAI's web interface.} to verbalize each relation class using natural language and placeholders for subject and object entities (see Appendix \ref{app:template_gen} for more details). The placeholders within the hypothesis templates are replaced by the entity types, $e_{1_{type}}$ and $e_{2_{type}}$, found in the premise. 

\textbf{Feasible Hypothesis Filter:}
There is an implicit multiplicative effect of adapting RE into an NLI task where each relationship instance produces $m$ class-indicative hypotheses resulting in $|X_{\textsc{RE}}| \times m$ premise-hypothesis pairs. To mitigate this effect, we develop a feasible hypothesis filter which automatically filters out improbable hypotheses by aggregating valid sets of entity-type pairs by relationship classes across all training data: $E_{valid}=\left\{r_1 \mapsto S_1, r_2 \mapsto S_2, \ldots, r_m \mapsto S_m\right\}$ where $r_j \in R$ for $j=1,2, \ldots, m$ and each $S_j$ is the set of tuples of entity-type pairs associated with all instances of relationship class $r_j$. 

Using this filter, we assess the feasibility of hypotheses given a pair of entity types: $\hat{H}_i = \{h_j | (e_{1_{type}}, e_{2_{type}}) \in E_{valid}(r_{j})  \}_{j=1}^{|R|}$ where $\hat{H}_i$ is a set of feasible hypotheses given the entity-type pair found in instance $i$, and  $\hat{H}_i \subset H$ where $H$ is the set of all possible hypotheses. 

Since sets of feasible hypotheses are approximated using the training data's relationships and entity-type pairs, the filter may remove valid hypotheses based on an entity-type pair and corresponding relation that exists only in the test set. For these instances, the entailed premise-hypothesis pair will not be presented to the model, leading to false negatives. However, in practice, we observe that this does not occur with the datasets we use for evaluation and should not occur as long as training data is sufficiently representative of the test data (i.e., the training data contains at least one relationship with a specific entity-type pair for every relation and entity-type pair found in the test set). 

\textbf{Meta-class Analysis (MCA): }  
After applying the aforementioned feasible hypothesis filter, we leverage MCA to assign NLI labels, namely \textit{entail}, \textit{neutral}, and \textit{contradict}, to the resultant premise-hypothesis pairs. To do this, we first construct definition-based mutually exclusive meta-relationships between relation classes. For example, in the ChemProt dataset, the ``up regulator'' class is, by definition, mutually exclusive to the ``down regulator'' class. For datasets with a negative class (e.g., ``no relation''), the negative class is mutually exclusive to all positive classes and vice-versa. With this analysis, we construct NLI labels in the following way: 

\begin{enumerate}[label=(\alph*),nosep,leftmargin=*]    
\item \textit{Entail}: Premise-hypothesis pairs are labeled ``entail'' when the hypothesis $h_j$ aligns with the verbalized ground truth label $y_i$. 
\item \textit{Neutral}: If the original instance expresses a positive class (i.e., any class other than the ``no relation'' class), then all non-exclusive class hypotheses are labeled as ``neutral.'' 
\item  \textit{Contradict}: The ``contradict'' label is assigned to hypotheses that verbalize definitionally exclusive classes.
\end{enumerate}
See Appendix \ref{app:metaclass} for tables showing how relation labels map to NLI labels using MCA.

\textbf{LLM Fine-tuning: } With generated premise-hypothesis pairs, we train a discriminative language model, namely BioLinkBERT\sst{large} \cite{linkbert}, to predict NLI labels. We concatenate premise-hypothesis pairs as the input to the language model and send the resultant representation of the special {\small \textsc{[CLS]}} token through a fully connected layer, which is trained using cross-entropy loss:
\vspace{-4mm}
\begin{equation}\label{eq:cross_entropy}
\begin{aligned}
\mathcal{L}_{\mathrm{CE}}=-\sum_{i=1}^{m} y_{o,i} \cdot \log \left(p\left(y_{o,i}\right)\right)
\end{aligned}
\vspace{-2mm}
\end{equation}
where $y$ is a binary indicator that is $1$ if and only if $i$ is the correct classification for observation $o$, $p(y_{o,i})$ is the softmax probability that observation $o$ is of class $i$, and $m$ is the number of classes.

\textbf{Group-based Prediction Selection:} Given the multiplicative effect of adapting RE-to-NLI where one relation instance results in a group of up to $m$ premise-hypothesis pairs, we can employ a group selection method to select the most confident \textit{entail} prediction. If the model predicts two or more entailed instances within a group of premise-hypothesis pairs, we use the softmax probability from Equation \ref{eq:cross_entropy} as a proxy for model confidence \cite{hendrycks17baseline} and select the prediction with the highest confidence. We allow the model to naturally abstain from making a prediction by predicting ``neutral'' for all premise-hypothesis pairs in a group.

\section{Experiments}\label{sec:experiments}

\subsection{Datasets}\label{sec:datasets}
We include a spread of experiments on various biomedical RE datasets. BioRED is a document-level RE dataset featuring eight relation classes \cite{biored}. BioRED also provides an orthogonal and binary ``Novel'' class, which annotates whether an instance expresses a novel finding. BC5CDR is a document-level RE dataset featuring binary relations between chemical and disease entities \cite{Li2016BioCreativeVC}. DDI13 is a drug-drug interaction dataset with four relation classes \cite{ddi13}, and ChemProt is a chemical-protein dataset featuring five relation classes \cite{Taboureau2010ChemProtAD}. GAD is a gene-disease dataset with binary relations \cite{gad}. We only include GAD in our main experiment for comparative purposes to past works. We believe that the GAD dataset should be retired from future works due to significant label accuracy issues, which the authors acknowledge.\footnote{\url{https://github.com/dmis-lab/biobert/issues/162}}

ChemProt and BioRED, unlike the other datasets in our experiments, do not annotate negative instances, leading to ambiguity in handling unannotated data. These unannotated instances can be addressed in three ways: (1) by treating them uniformly as a negative class, (2) by considering them as potential members of novel, unannotated classes, or (3) by using a generalized approach that considers unannotated instances as a mix of negative and novel classes. To avoid this subjectivity, we focus only on annotated instances for training and evaluation, applying this approach consistently across all datasets to ensure a fair comparison of methods.

As mentioned in Section \ref{sec:introduction}, our method is designed to leverage features of biomedical domain RE, namely the prevalence of definitionally exclusive classes and the importance of entity types vis-à-vis feasible relationships. However, we also seek to assess our method beyond the biomedical domain and extend our experiments to general domain datasets ReTACRED \cite{ReTACREDAS} and SemEval-2010 Task 8 \cite{Hendrickx2010SemEval2010T8}. ReTACRED is a re-annotated version of TACRED \cite{zhang2017tacred} and features 40 relation classes---significantly more classes than any of the biomedical datasets we tested. SemEval-2010 is a sentence-level RE dataset with ten relation classes.

\subsection{Baselines}
\subsubsection{Traditional Multi-Class Classification}
We select leading biomedical language models and train them using a traditional RE multi-class approach where models directly predict relation classes. \textbf{BioM-ALBERT\sst{xxlarge}} and \textbf{BioM-BERT\sst{large}} \cite{biom} are transformer architectures adapted into the biomedical domain by using a custom biomedical vocabulary and pre-training on PubMed abstracts \cite{pubmed} and PubMed Central articles \cite{pmc}. \textbf{BioMed RoBERTa\sst{base}} \cite{biomed_roberta} features the RoBERTa architecture \cite{liu2019roberta} adapted to the biomedical domain via continued pre-training on papers from the S2OR Corpus \cite{Lo2020S2ORCTS}. \textbf{PubMedBERT\sst{base}} \cite{pubmedbert} and \textbf{BioLinkBERT\sst{large}} \cite{linkbert} are BERT \cite{devlin-etal-2019-bert} variants. The former is trained on PubMed abstracts with a custom biomedical vocabulary. The latter is trained with two self-supervised objectives: masked language modeling and document relation prediction. 

\subsubsection{NLI Adapted Models}
\textbf{NBR} is a biomedical domain RE-to-NLI method that leverages BioLinkBERT\sst{large} as a backbone language model. Like our method, NBR converts relation instances and labels into premise-hypothesis pairs. Key differences between NBR and our method are that NBR does not use MCA or feasible hypothesis filtering, and they leverage a ranking loss training objective to rank entailed premise-hypothesis pairs over non-entailed pairs. 

The RE-to-NLI adaptation method used in \our is architecture-agnostic, so we also experiment with auto-regressive architectures. We conduct the following experiment using identical data and methods to those discussed in Section \ref{sec:methods}; the only difference is the final training step.

We fine-tune \textbf{Phi-2} (2.7B) and \textbf{Phi-3} (3.8B).\footnote{We use the \textit{microsoft/phi-2} and \textit{microsoft/Phi-3-mini-4k-instruct} checkpoints from Hugging Face.} For Phi-2 and Phi-3, we construct a seq-to-seq task and fine-tune the models to generate an NLI label for each premise-hypothesis pair. For more information about training Phi-2 and Phi-3, see Appendix \ref{app:phi2phi3}.

We also seek to assess the performance of large, frontier auto-regressive language models, \textbf{GPT 3.5} \cite{gpt35} and \textbf{GPT 4} \cite{gpt4},\footnote{Specifically, we use \textit{gpt-3.5-turbo-0125} and \textit{gpt-4-turbo-2024-04-09} via OpenAI's API.} leveraging few-shot, in-context learning. For more on the prompts we use to solicit predictions from GPT 3.5 and GPT 4, see Appendix \ref{app:gpt35gpt4}.

For all NLI-adapted models, only entailed premise-hypothesis pairs map directly to the original RE training instance. Thus, we only keep NLI instances labeled or predicted as entailed when mapping instances back into the original RE labels for evaluation. This ensures a fair comparison across adapted and non-adapted methods.

\subsection{General Domain Experiments}
For our general domain experiments, we use \textbf{DeBERTaV3\sst{large}} \cite{debtera_v3} and \textbf{RoBERTa-MNLI\sst{large}} \cite{liu2019roberta}. DeBERTaV3 is an improved version of BERT that uses replaced token detection, a more sample-efficient pre-training objective. RoBERTa-MNLI\sst{large} is the RoBERTa architecture fine-tuned on the MNLI corpus \cite{mnli}.\footnote{We use the \textit{FacebookAI/roberta-large-mnli} checkpoint from Hugging Face.} 

We make slight modifications to the general domain version of \our. We use RoBERTa-MNLI\sst{large} as the backbone language model, and we do not leverage surface-form abstraction for entity types (i.e., we leave the original entities as they appear in the text and do not replace them with their corresponding types). Entity surface form abstraction is a method developed for the long-tail nature of biomedical entities \cite{Peng2020LearningFC}. Also, some general domain RE datasets, such as SemEval-2010 Task 8, do not provide annotated entity type information.

\section{Results}\label{sec:results}

\begin{table*}[!htbp]
\centering
\begin{adjustbox}{max width=1.0\textwidth}
\begin{tabular}{ l | c | c | c | c | c | c } 
    \bb{Model}	&	\bb{BC5CDR}	&	\bb{BioRED}	&	\bb{BioRED (novel)}	&	\bb{ChemProt}	& \bb{DDI13}	&	\bb{GAD}	\\ \specialrule{1.2pt}{0pt}{1pt}
    \multicolumn{7}{c}{\textsc{Traditional Multi-class Classification}}  \\ \hline
    BioM-ALBERT\sst{xxlarge} \cite{biom}  &	0.679	&	0.668	&	0.863	&\uu{0.940}	&	0.911	&	0.815	\\
    BioM-BERT\sst{large} \cite{biom}  &	0.681	&   0.709	& \uu{0.904}	&	0.934	&	\uu{0.917}	&	0.795	\\
    BioMed RoBERTa\sst{base} \cite{biomed_roberta}	&	0.664	&	0.714	&	0.897	&	0.919	&	0.911	&	0.803	\\
    PubMedBERT\sst{base} \cite{pubmedbert}	    &	0.651	&\uu{0.715} &	0.891	&	0.923	&	0.916	&	0.803	\\
    BioLinkBERT\sst{large} \cite{linkbert}	&	0.682	&	0.699	&	0.899	&	0.931	&\uu{0.917}	&	0.806	\\ \hline \hline
    \multicolumn{7}{c}{\textsc{NLI Adapted Models}}  \\ \hline
    NBR \cite{nbr}&	0.679	&	0.543	&	0.664	&	0.883	&	0.846	&	\uu{0.831}	\\
    Phi-2 \cite{textbooks2} &	0.653	&\uu{0.715} &	0.824	&	0.852	&	0.873	&	0.729	\\
    Phi-3 \cite{phi3}       & \uu{0.749}       &  0.688       & 0.840       & 0.930        &   0.915   & 0.721 \\
    GPT 3.5\textsuperscript{$\dagger$}	\cite{gpt35}        &	0.282	&	0.470	&	0.594	&	0.494	&	0.386	&	0.548	\\
    GPT 4\textsuperscript{$\dagger$} \cite{gpt4}          &	0.418	&	0.532	&	0.680	&	0.626	&	0.492	&	0.660	\\ \hline
    \bb{\our}   &	\bb{0.757}	&	\bb{0.891}	&	\bb{0.917}	&	\bb{0.968}	&	\bb{0.957}	&	\bb{0.878}	\\
    \specialrule{1.2pt}{0pt}{1pt}
\end{tabular}\end{adjustbox} 
\caption{Micro F1 scores for traditional RE and NLI adapted methods. $\dagger$Results from GPT 3.5 and GPT 4 are via in-context learning (see Appendix \ref{app:gpt35gpt4} for details), whereas other models were fine-tuned directly on the task from our own implementations. Results show averages over five runs.}
\label{tab:main_results}
\vspace{\inteval{\y}mm}
\end{table*}

We observe an interesting comparison between the BioLinkBERT\sst{large} model and \our. Both experiments share the same backbone language model, yet the performance of our \our method is significantly higher providing evidence of the effectiveness of adapting the RE task into one of textual entailment. We hypothesize that the boost in performance primarily comes from the additional data abstraction RE-to-NLI introduces by training the model to recognize entailed premise-hypothesis pairs instead of directly predicting suppositional classes. By combining RE-to-NLI adaptation with surface-form entity abstraction, the model is less prone to memorizing entities and shallow heuristics of relation classes; instead, it must understand the context and the natural language interplay between a premise and hypothesis. Furthermore, the boost in performance between the NBR model and \our highlights the effectiveness of leveraging MCA, feasible hypothesis filtering, and group-based prediction selection.

Within the biomedical domain experiments, the NLI-adapted auto-regressive models generally underperform compared to the discriminative models. Predictably, the larger Phi-3 outperforms Phi-2 and fine-tuning smaller auto-regressive models outperforms larger models, GPT 3.5 and GPT 4, leveraging few-shot in-context learning. This aligns with findings from \citet{peng2024metaie} that LLMs using in-context learning underperform relative to smaller, fine-tuned language models on information extraction tasks. 

We observe better performance from auto-regressive architectures in the general domain. The performance from Phi-3 approaches that of \our on both ReTACRED and SemEval-2010 Task 8 datasets which is promising for auto-regressive models, in general. We leave fine-tuning larger auto-regressive models to future work but expect additional gains to be made, potentially overtaking the discriminative models.

\begin{table}[t]
\begin{adjustbox}{max width=.48\textwidth}
\begin{tabular}{ l | c | c } 
    \bb{Model}	&	\bb{ReTACRED}	&	\bb{SemEval}	\\ \specialrule{1.2pt}{0pt}{1pt}
    \multicolumn{3}{c}{\textsc{Traditional Multi-class Classification}}  \\ \hline
    DeBERTaV3\sst{large} \cite{debtera_v3} & 0.809 & 0.807 \\
    RoBERTa-MNLI\sst{large} \cite{liu2019roberta} &	0.800	&	0.828	\\ \hline \hline
    \multicolumn{3}{c}{\textsc{NLI Adapted Models}}  \\ \hline
    NBR \cite{nbr}	&	0.875	&	0.826	\\
    Phi-2 \cite{textbooks2}	&	0.862	&	0.855	\\
    Phi-3 \cite{phi3}	&	\uu{0.880}	&	\uu{0.871}	\\ 
    GPT 3.5\textsuperscript{$\dagger$} \cite{gpt35}	&	0.306	&	0.340	\\
    GPT 4\textsuperscript{$\dagger$} \cite{gpt4}	&	0.565	&	0.616	\\ \hline
    \textbf{\our}	&	\textbf{0.943}	&	\textbf{0.902}	\\
\specialrule{1.2pt}{0pt}{1pt}
\end{tabular}\end{adjustbox} 
\caption{Micro F1 scores from general domain RE experiments.}
\label{tab:gen_domain_results}
\vspace{\inteval{\y}mm}
\end{table}

\subsection{Ablation Experiments}\label{sec:ablations}

\begin{table}[t]
\begin{adjustbox}{max width=.48\textwidth}
\begin{tabular}{ l | c | c | c } 
    \bb{Model}	&	\bb{BioRED}	&	\bb{ChemProt} &	\bb{ReTACRED}	\\ \specialrule{1.2pt}{0pt}{1pt}
    \our	&	0.891	&	0.968 &	0.943	\\ \hline
    \quad (w/o Feasible Hypothesis Filter)	&	0.876	&	N/A &	DNC	\\
    \quad (w/o Meta-class Analysis) 	&	0.853	&	0.911 &	0.916	\\
    \quad (w/o Grouped Selection)	&	0.805	& 0.950 &	0.875	\\
\specialrule{1.2pt}{0pt}{1pt}
\end{tabular}\end{adjustbox} 
\caption{Micro F1 scores from ablation experiments which remove each proposed module within \our. Each module has a significant impact on performance. ChemProt is monolithic in its entity types (\textit{chemicals} and \textit{diseases}), which prevents the use of the feasible hypothesis filter. On ReTACRED, we observe that without applying the feasible hypothesis filter, the model does not converge (DNC).}
\label{tab:ablation_results}
\vspace{\inteval{\y}mm}
\end{table}

We conduct ablations to better understand \our's performance gains by removing modules and reporting the performance. Note that the performance of BioLinkBERT\sst{large} in Table \ref{tab:main_results} can be considered an ablation of \our that does not leverage NLI adaptation or any additional modules since \our uses the same backbone language model. For our ablations, we choose to examine the BioRED, ChemProt, and ReTACRED datasets because they feature more than two relation classes and contain one or more definition-based mutually exclusive relations as determined by MCA.

\begin{figure*}[th!]    
    \includegraphics[width=1.0\textwidth, keepaspectratio]{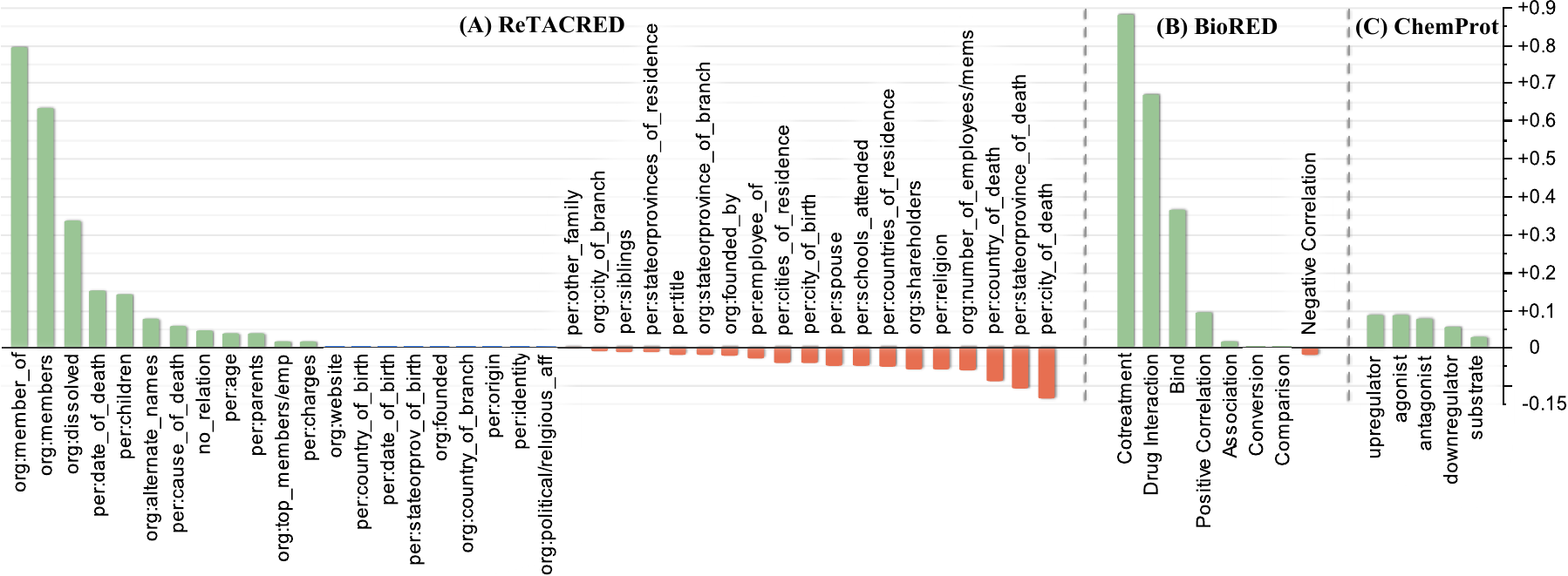}
    \vspace*{-4mm}
    \caption{$\Delta$\textsc{F1} per relation class when leveraging meta-class analysis to assign NLI labels.} \label{fig:cs_combo}
    \vspace{-4mm}
\end{figure*}

\begin{enumerate}[label=(\alph*),nosep,leftmargin=*]    
\item \textit{w/o Feasible Hypothesis Filter}: We remove the feasible hypothesis filter, and, in doing so, each relation instance is converted into $m$ premise-hypothesis pairs, with $m$ being the number of classes in a dataset. This produced a moderate drop in performance on BioRED ($m=8$). Since the feasible hypothesis filter is based on entity type pairs, it is not available (N/A) for datasets such as ChemProt, which only feature a single entity type pair (namely, \textit{chemical} and \textit{gene}) associated with every relation class. However, the feasible hypothesis filter is essential in model convergence when a dataset consists of many relation classes, such as ReTACRED ($m=40$). The model did not converge on ReTACRED without the feasible hypothesis filter, likely due to the overwhelming number of non-informative ``neutral'' premise-hypothesis pairs used in training.

\item \textit{w/o Meta-class Analysis}: Removing MCA and using ``neutral'' as the NLI label for all non-entailed premise-hypothesis pairs led to a considerable drop in performance, indicating the benefit of training the model with the additional training signal obtained via MCA. Note that in this ablation experiment, we maintain mutual exclusive NLI labels between positive and negative (i.e., ``no relation'') classes.

\item \textit{w/o Group Prediction Selection}: Without this module, we select \textit{all} entailed predictions regardless of how many \textit{entail} predictions are made within a group of premise-hypothesis pairs. Doing this allows the model to predict multiple classes for a single relation instance. This ablation experiment led to a drop in performance across all datasets but most significantly on BioRED, which we suspect results from the closeness in BioRED's ``positively correlated'' and ``associated'' relation classes, as ``associated'' can sometimes be considered a hypernym of ``positively correlated,'' leading the model to predict \textit{entail} for both of the corresponding hypotheses.
\end{enumerate}

\subsection{Meta-class Analysis Case Study}

To further explore the impacts of leveraging MCA, we decompose results from ReTACRED, BioRED, and ChemProt by evaluating the change in Micro F1 scores ($\Delta$F1) for each class. We isolate the effect of MCA by training identical models with and without MCA-informed NLI labels and report the results in Figure \ref{fig:cs_combo}.

MCA results in a net benefit in performance across classes and datasets, but the specific nature of these benefits varies. In ReTACRED, we observe notable improvements in the ``member of'' and ``members'' classes, which are definitionally exclusive. Conversely, some classes experience minor decreases in performance. For BioRED, we observe a slight drop in predictive performance for the ``negative correlation'' class, while all other classes get a significant boost. The largest performance gains are seen in classes that are not mutually exclusive, suggesting that the additional training signal from MCA aids the model in disentangling adjacent relation class representations. In ChemProt, we observe near-uniform, albeit relatively small, boosts in performance across all classes. This indicates that MCA has varied effects across disparate datasets. There is a net benefit but, interestingly, the exact nature of the benefit varies across datasets.

\subsection{Additional Experiments}
Given that RE-to-NLI adaptation leads to models predicting the same \textit{entail}, \textit{neutral}, \textit{contradict} labels across disparate datasets, we naturally sought to investigate the potential of combining the relatively small and disjoint biomedical RE datasets into a single, unified task. Unfortunately, these experiments failed to produce significant performance gains, indicating that these biomedical datasets have limited synergistic effects when adapted to the NLI task (see Appendix \ref{app:task_unification}).

\section{Conclusion}\label{sec:conclusion}
The exploration of NLI techniques to enhance relation extraction has opened new avenues in natural language processing, and our study introduces \our as an advancement in this area. By adapting the RE task into an NLI framework and incorporating innovative strategies such as meta-class analysis, feasible hypothesis filtering, and group-based prediction selection, \our demonstrates remarkable improvements in RE performance. Our experiments, conducted across biomedical and general domain datasets, highlight the robustness and versatility of \our. By openly sharing our code, experimental settings, and datasets, we aim to facilitate further research and development in this promising intersection of NLI and RE, paving the way for more sophisticated and accurate information extraction systems in diverse domains.



\section*{Limitations}
\our is not without its limitations. By verbalizing a hypothesis for each relation class, the training data is multiplied by the number of relation classes in the dataset, necessitating additional training resources. Our introduced module, the feasible hypothesis filter, relies heavily on accurate entity-type information. This information is crucial for the success of the adaptation process. However, the filtering process becomes ineffective if this information is unavailable or if numerous feasible hypotheses (e.g., 40+) exist for a given relation class and entity type pair. In these scenarios, the ``entail'' class becomes a minority class in a sea of ``neutral'' NLI instances, potentially causing the model to collapse to a trivial state of simply predicting ``neutral'' for every premise-hypothesis pair. Such a scenario would require the design of manually tuned sampling strategies or bespoke learning objectives to handle the overwhelming number of ``neutral'' premise-hypothesis pairs. We defer the exploration of such challenging settings to future research.

Additionally, in our study, meta-class analysis is performed manually, which introduces an extra layer of human effort. This manual effort involves reading annotation guidelines for a specific dataset to determine which relation classes are mutually exclusive based on their definitions. While this task is relatively quick and straightforward, it does require additional human involvement.

\section*{Ethics Statement}
We do not anticipate any major ethical concerns; relation extraction is a fundamental problem in natural language processing. A minor consideration is the potential for introducing certain hidden biases into our results (i.e.,~performance regressions for some subset of the data despite overall performance gains). However, we did not observe any such issues in our experiments, and indeed these considerations seem low-risk for the specific datasets studied here because they are all published. 

\bibliography{anthology,custom}

\appendix
\section{Appendix} \label{sec:appendix}

\subsection{Automatic Generation of Hypothesis Templates}\label{app:template_gen}
To reduce human effort in our methods, we turn to LLMs, specifically GPT 3.5 \cite{gpt35}, to automatically generate hypothesis templates. Some datasets, such as BC5CDR, GAD, and BioRED Novel, feature two classes, making the template generation process relatively trivial. The benefits of automating the generation of hypothesis templates are more significant for datasets such as ReTACRED, which feature 40 relation classes. 

We use the following prompt where the ellipsis is replaced with the list of natural language relation classes (e.g., relation classes with underscores removed and spaces inserted) used in each dataset:

\begin{blockquotec}
Verbalize the following relation classes in the form ``subj [verbalized relation] obj": [\ldots].
\end{blockquotec}

A special case arose for the DDI13 dataset where each relation instance describes a relation between two drugs. We referenced the verbalized hypotheses proposed by \citet{nbr} and included instructions about describing two drug entities:

\begin{blockquotec}
Verbalize the following relation classes using the form "[verbalized relation] two drugs is described": [\ldots].
\end{blockquotec}

Table \ref{tab:hypo_templates} contains the generated hypothesis templates for each dataset.

\begin{table*}[th!]
\centering
\begin{adjustbox}{max width=1.0\textwidth,totalheight=0.90\textheight,keepaspectratio}

\begin{tabular}{| l | l | l |} \hline

\bb{Dataset} &	\bb{Relation Classes} & \bb{Hypothesis Templates}	\\ \specialrule{1.2pt}{0pt}{0pt}
BC5CDR & Associated & ``subj is associated with obj.''\\
& Not Associated & ``subj is not associated with obj.''\\ \hline
BioRED  & Positive Correlation&  ``subj positively correlates with obj.'' \\
& Negative Correlation&  ``subj negatively correlates with obj.''\\
& Association&  ``subj is associated with obj.''\\
& Comparison&  ``subj is compared with obj.''\\
& Conversion&  ``subj converts to obj.''\\
& Cotreatment&  ``subj is co-treated with obj.''\\
& Drug Interaction &  ``subj interacts with obj (as drugs).''\\
& Bind &  ``subj binds to obj.''    \\ \hline
BioRED Novel & Novel & ``subj introduces a novel relationship to obj.'' \\
& Not novel & ``subj does not introduce a novel relation to obj.''\\ \hline
ChemProt & Upregulator & ``subj upregulates obj.''\\
& Downregulator & ``subj downregulates obj.''\\
& Agonist & ``subj acts as an agonist for obj.''\\
& Antagonist & ``subj acts as an antagonist for obj.''\\
& Substrate & ``subj is a substrate for obj.''\\ \hline
DDI13 & Advise & "Advice regarding two drugs is described.''\\
& Effect & "An effect between two drugs is described.''\\
& Interaction & "An interaction between two drugs is described.''\\
& Mechanism & "The mechanism involving two drugs is described.''\\\hline
GAD & Associated & ``subj is associated with obj.''\\
& Not Associated & ``subj is not associated with obj.''\\ \hline
ReTACRED & No relation & ``subj has no relation with obj.''\\
& Org:alternate names & ``subj has alternate names as obj.''\\
& Org:city of branch & ``subj's branch is located in the city of obj.''\\
& Org:country of branch & ``subj's branch is located in the country of obj.''\\
& Org:dissolved & ``subj has been dissolved.''\\
& Org:founded & ``subj was founded on the date obj.''\\
& Org:founded by & ``subj was founded by obj.''\\
& Org:member of & ``subj is a member of obj.''\\
& Org:members & ``subj has members including obj.''\\
& Org:number of employees/members & ``subj has obj number of employees/members.''\\
& Org:political/religious affiliation & ``subj has political/religious affiliation with obj.''\\
& Org:shareholders & ``subj has shareholders including obj.''\\
& Org:state or province of branch & ``subj's branch is located in the state or province of obj.''\\
& Org:top members/employees & ``subj's top members/employees include obj.''\\
& Org:website & ``subj's website is obj.''\\
& Per:age & ``subj's age is obj.''\\
& Per:cause of death & ``subj's cause of death is obj.''\\
& Per:charges & ``subj is charged with obj.''\\
& Per:children & ``subj has obj as children.''\\
& Per:cities of residence & ``subj resides in cities including obj.''\\
& Per:city of birth & ``subj was born in the city of obj.''\\
& Per:city of death & ``subj died in the city of obj.''\\
& Per:countries of residence & ``subj resides in countries including obj.''\\
& Per:country of birth & ``subj was born in the country of obj.''\\
& Per:country of death & ``subj died in the country of obj.''\\
& Per:date of birth & ``subj was born on the date obj.''\\
& Per:date of death & ``subj died on the date obj.''\\
& Per:employee of & ``subj is an employee of obj.''\\
& Per:identity & ``subj's identity is obj.''\\
& Per:origin & ``subj's origin is obj.''\\
& Per:other family & ``subj has obj as other family members.''\\
& Per:parents & ``subj's parents include obj.''\\
& Per:religion & ``subj's religion is obj.''\\
& Per:schools attended & ``subj attended schools including obj.''\\
& Per:siblings & ``subj's siblings include obj.''\\
& Per:spouse & ``subj's spouse is obj.''\\
& Per:state or province of birth & ``subj was born in the state or province of obj.''\\
& Per:state or province of death & ``subj died in the state or province of obj.''\\
& Per:state or provinces of residence & ``subj resides in states or provinces including obj.''\\
& Per:title & ``subj's title is obj.''\\ \hline
SemEval 2010 & Other & ``subj and obj are related in some other way.''\\
& Component-Whole & ``subj is a component of obj.''\\
& Instrument-Agency & ``subj is used by obj.''\\
& Member-Collection & ``subj is a member of obj.''\\
& Cause-Effect & ``subj causes obj.''\\
& Entity-Destination & ``subj is taken to obj.''\\
& Message-Topic & ``subj is about obj.''\\
& Entity-Origin & ``subj comes from obj.''\\
& Product-Producer & ``subj is produced by obj.''\\
& Content-Container & ``subj contains obj.'' \\
\specialrule{1.2pt}{0pt}{1pt}
\end{tabular}
\end{adjustbox} 
\caption{Auto-generated hypothesis templates for each relation class in each dataset. Hypotheses are generated using GPT 3.5 and the prompt described in Appendix \ref{app:template_gen}.}
\label{tab:hypo_templates}
\end{table*}

\subsection{Baselines}\label{app:baselines}
\subsubsection{GPU Resources} 
All baselines were trained on a single NVIDIA A100, and training times ranged from 1 to 12 hours, changing based on the size of the dataset and the number of parameters in the model.  

\subsubsection{Phi-2 and Phi-3}\label{app:phi2phi3}
Since responses from auto-regressive models may sometimes include additional text, all responses are aligned to ground truth labels using partial string matching. We do this by searching for the matches of the first three letters in each NLI label (e.g., ``ent'' $\rightarrow$ \textit{entail}, ``con'' $\rightarrow$ \textit{contradict}, ``neu'' $\rightarrow$ \textit{neutral}). When a class cannot be matched, we assign ``none,'' which, during evaluation, is equivalent to the NLI label \textit{neutral}.

For Phi-2, we use the following prompt to fine-tune the model on our task:
\begin{blockquotec}
[INST]You are given a premise and a hypothesis below. If the premise entails the 
hypothesis, return ``entail.'' If the premise contradicts the hypothesis, 
return ``contradict.'' Otherwise, if the premise does neither, return ``neutral.''[/INST]\\
\\
\#\#\# Premise: [premise]\\
\#\#\# Hypothesis: [hypothesis]\\
\#\#\# Label: [nli\_target]
\end{blockquotec}
\noindent Phi-3 uses a similar prompt that differs only in format:
\begin{blockquotec}
<|system|>\\
You are given a premise and a hypothesis below. If the premise entails the 
hypothesis, return ``entail.'' If the premise contradicts the hypothesis, 
return ``contradict.'' Otherwise, if the premise does neither, return ``neutral.''\\
<|end|>\\
\\
<|user|> \\
Premise: [premise]\\
Hypothesis: [hypothesis]\\
Label:\\
<|end|>\\
\\
<|assistant|>

[nli\_target] \\ 
<|end|>
\end{blockquotec}

Both Phi-2 and Phi-3 were fine-tuned using the hyperparameters in Table \ref{tab:phi2_phi3_hp}.

\begin{table}[ht!]
\begin{adjustbox}{max width=.48\textwidth} 
\begin{tabular}{| l | c |}
\hline
\bb{Parameter}	&	\bb{Value}	\\ \specialrule{1.2pt}{0pt}{0pt}
Epochs & 3 \\
Max seq. length & 1,024 \\ 
Batch size & 3 \\
Grad. accumulation steps & 2 \\
Max gradient norm & 0.3 \\
Learning rate & 2e-4\\ 
Lr scheduler type & cosine \\
Weight decay & 0.001 \\
Warm-up ratio & 0.03 \\ 
\specialrule{1.2pt}{0pt}{0pt}
\end{tabular}\end{adjustbox} 
\caption{Hyperparameters used to fine-tune Phi-2 and Phi-3.}
\label{tab:phi2_phi3_hp}
\end{table}

\subsubsection{GPT 3.5 and GPT 4}\label{app:gpt35gpt4}
GPT 3.5 and GPT 4 often perform better on tasks with the help of in-context learning~\cite{Wei2023LargerLM, wang2023large}. We construct a prompt that lists the NLI labels and offers four examples of premise-hypothesis pairs expressing each NLI label. 

The following is the prompt we used for soliciting predictions for our tests:

\begin{blockquotec}
You are given a premise and a hypothesis below. If the premise entails the hypothesis, return ``entail.'' If the premise contradicts the hypothesis, return ``contradict.'' Otherwise, if the premise does neither, return ``neutral.'' The following are some examples:
\\
\\
\#\#\# Premise: [premise]\\
\#\#\# Hypothesis: [hypothesis]\\
\#\#\# Label: [nli\_target]
\\ ...\\ \{\textit{4x examples of each NLI class are provided}\}\\...\\
\end{blockquotec}

For responses from GPT 3.5 and GPT 4, we use the same partial string matching used for Phi-2 and Phi-3 (Appendix \ref{app:phi2phi3}) for evaluation. 

\subsubsection{Hyperparameters for \our}
Table \ref{tab:meta_entail_re_hyperparams} contains the hyperparameters used to train \our. 

\begin{table}[ht]
\begin{adjustbox}{max width=.48\textwidth} 
\begin{tabular}{| l | c |}
\hline
\bb{Parameter}	&	\bb{Value}	\\ \specialrule{1.2pt}{0pt}{0pt}
Epochs & 3 \\
Batch size & 32 \\
Grad. accumulation steps & 1 \\
Max seq. length & 1,024 \\ 
Learning rate & 2e-5 \\
Seeds & \{41, 42, 43, 44, 45\} \\
Optimizer & AdamW \\
LR Scheduler warm-up steps & 0 \\
LR Scheduler training steps & 1,000 \\
\specialrule{1.2pt}{0pt}{0pt}
\end{tabular}\end{adjustbox} 
\caption{Hyperparameters used to fine-tune BioLinkBERT\sst{large} for the \our method.}
\label{tab:meta_entail_re_hyperparams}
\end{table}

\subsection{Task Unification Results}\label{app:task_unification}
We explore unifying biomedical relation extraction datasets in hopes of boosting performance on a target dataset. We investigate two task-unification training methodologies: single-stage training and double-stage training. Single-stage training can be viewed as multi-task learning, where the model is trained simultaneously on multiple datasets and tested on a target dataset. Double-stage training can be viewed as an initial pre-training stage on all data except the target dataset, followed by fine-tuning and evaluation on the target dataset.

Unfortunately, we did not observe a significant performance boost across our task-unification experiments (see Table \ref{tab:task_unification}), potentially indicating that these biomedical datasets do not provide complementary information when adapted into an NLI task. Generally, the two-stage training is more effective than the single-stage training, but both fail to realize significant performance gains on the target datasets. We leave investigating other task-unification methods for future works.


\begin{table*}[t]
\centering
\begin{adjustbox}{max width=1.0\textwidth}
\begin{tabular}{ l l l | c } 

\multicolumn{4}{c}{\textsc{\textbf{Multi-task Learning (single-stage)}}}  \\ \specialrule{1.2pt}{0pt}{1pt}
\multicolumn{2}{l}{\textsc{Ensemble training data }} $\rightarrow$ 	 &	\textsc{Test set}	&	$\Delta$~\textsc{F1} \\ \hline 
\multicolumn{2}{l}{BC5CDR + BioRED + ChemProt + DDI13}	&	BC5CDR	&	-0.049 \\
\multicolumn{2}{l}{BioRED + ChemProt + DDI13 + BC5CDR}	&	BioRED	&	-0.031 \\
\multicolumn{2}{l}{ChemProt + BioRED + DDI13 + BC5CDR}	&	ChemProt	&	+0.012 \\ 
\multicolumn{2}{l}{DDI13 + BioRED + ChemProt + BC5CDR}	&	DDI13	&	-0.007 \\ \hline 
\\
\multicolumn{4}{c}{\textsc{\textbf{Continued pre-training with supervised fine-tuning (double-stage)}}}  \\ \specialrule{1.2pt}{0pt}{1pt}
\multicolumn{1}{l}{\textsc{Pre-training corpus}}	$\rightarrow$ &	\multicolumn{1}{l}{\textsc{Fine-tuning}}	$\rightarrow$ &	\textsc{Test set}	&	$\Delta$\textsc{F1} \\ \hline 
BioRED + ChemProt + DDI13	&	BC5CDR	&	BC5CDR	&	-0.005 \\
ChemProt + DDI13 + BC5CDR	&	BioRED	&	BioRED	&	-0.014 \\
BioRED + DDI13 + BC5CDR	&	ChemProt	&	ChemProt	&	-0.008 \\
BioRED + ChemProt + BC5CDR	&	DDI13	&	DDI13	&	-0.011 \\ \hline

\end{tabular}\end{adjustbox} 
\caption{Results from single-stage and double-stage task unification experiments. $\Delta$F1 scores are relative to \our scores from Table \ref{tab:main_results}. We do not observe signification performance improvements from our task unification experiments and leave further experimentation to future work.}
\label{tab:task_unification}
\end{table*}
\subsection{Meta-class analysis}\label{app:metaclass}
We conduct a meta-class analysis for each dataset used in Section \ref{sec:experiments}. We leverage class definitions to determine sets of mutually exclusive classes. The following tables show how meta-class analysis converts original RE labels (row headers) into NLI labels. The $h(class)$ column headers denote verbalized hypotheses using the corresponding class. For each table, we use the following denote NLI labels: 
\begin{itemize}[nosep]   
    \item \colorbox{red!25}{0} $\rightarrow$ \textit{contradict}
    \item \colorbox{yellow!25}{1} $\rightarrow$ \textit{neutral}
    \item \colorbox{green!25}{2} $\rightarrow$ \textit{entail}
\end{itemize}

%
%
\begin{table}[t]
\begin{center}
\begin{adjustbox}{max width=.48\textwidth}
\begin{tabular}{ l | c | c |} \cline{2-3}
    &	h(Associated) & h(Not Associated) \\ \hline
    \multicolumn{1}{|l|}{Associated}          & \cc{green!25}2 & \cc{red!25}0 \\ \hline
    \multicolumn{1}{|l|}{Not Associated}      & \cc{red!25}0 & \cc{green!25}2 \\ \hline
\end{tabular}\end{adjustbox} 
\caption{Meta-class analysis for BC5CDR. The ``Associated'' class is definitionally mutually exclusive to the ``Not Associated'' class.}
\label{tab:mca_bc5}
\end{center}
\end{table}

%
%
\begin{table}[t]
\begin{center}
\begin{adjustbox}{max width=.48\textwidth}
\begin{tabular}{ l | c | c | c | c | c | c | c | c |} \cline{2-9}
                                        & \rb{90}{h(Positive Correlation)} & \rb{90}{h(Negative Correlation) } & \rb{90}{h(Association)} & \rb{90}{h(Comparison)} & \rb{90}{h(Conversion)} & \rb{90}{h(Co-treatment)} & \rb{90}{h(Drug Interaction)} & \rb{90}{h(Bind)}  \\ \hline
    \multicolumn{1}{|l|}{Positive Correlation}  & \cc{green!25}2 & \cc{red!25}0 & \cc{yellow!25}1 & \cc{yellow!25}1  & \cc{yellow!25}1 & \cc{yellow!25}1 & \cc{yellow!25}1 & \cc{yellow!25}1 \\ \hline
    \multicolumn{1}{|l|}{Negative Correlation}& \cc{red!25}0  & \cc{green!25}2 & \cc{yellow!25}1 & \cc{yellow!25}1 & \cc{yellow!25}1 & \cc{yellow!25}1 & \cc{yellow!25}1 & \cc{yellow!25}1\\ \hline
    \multicolumn{1}{|l|}{Association}       & \cc{yellow!25}1  & \cc{yellow!25}1 & \cc{green!25}2 & \cc{yellow!25}1 & \cc{yellow!25}1 & \cc{yellow!25}1 & \cc{yellow!25}1 & \cc{yellow!25}1\\ \hline
    \multicolumn{1}{|l|}{Comparison}    & \cc{yellow!25}1  & \cc{yellow!25}1 & \cc{yellow!25}1 & \cc{green!25}2 & \cc{yellow!25}1 & \cc{yellow!25}1 & \cc{yellow!25}1 & \cc{yellow!25}1 \\ \hline
    \multicolumn{1}{|l|}{Conversion}     & \cc{yellow!25}1  & \cc{yellow!25}1 & \cc{yellow!25}1 & \cc{yellow!25}1 & \cc{green!25}2  & \cc{yellow!25}1 & \cc{yellow!25}1 & \cc{yellow!25}1\\ \hline
    \multicolumn{1}{|l|}{Co-treatment}     & \cc{yellow!25}1  & \cc{yellow!25}1 & \cc{yellow!25}1 & \cc{yellow!25}1 & \cc{yellow!25}1 & \cc{green!25}2 & \cc{yellow!25}1 & \cc{yellow!25}1 \\ \hline
    \multicolumn{1}{|l|}{Drug Interaction}     & \cc{yellow!25}1  & \cc{yellow!25}1 & \cc{yellow!25}1 & \cc{yellow!25}1 & \cc{yellow!25}1 & \cc{yellow!25}1  & \cc{green!25}2 & \cc{yellow!25}1 \\ \hline
    \multicolumn{1}{|l|}{Bind}     & \cc{yellow!25}1 & \cc{yellow!25}1 & \cc{yellow!25}1 & \cc{yellow!25}1  & \cc{yellow!25}1 & \cc{yellow!25}1 & \cc{yellow!25}1 & \cc{green!25}2  \\ \hline
\end{tabular}\end{adjustbox} 
\caption{Meta-class analysis for BioRED. The ``Positive Correlation'' class is mutually exclusive to the ``Negative Correlation'' class.}
\label{tab:mca_biored}
\end{center}
\end{table}

%
%
\begin{table}[t]
\begin{center}
\begin{adjustbox}{max width=.48\textwidth}
\begin{tabular}{ l | c | c |} \cline{2-3}
    &	h(Novel) & h(Not Novel) \\ \hline
    \multicolumn{1}{|l|}{Novel}          & \cc{green!25}2 & \cc{red!25}0 \\ \hline
    \multicolumn{1}{|l|}{Not Novel}      & \cc{red!25}0 & \cc{green!25}2 \\ \hline
\end{tabular}\end{adjustbox} 
\caption{Meta-class analysis for BioRED Novel. The ``Novel'' class is mutually exclusive to the ``Not Novel'' class.}
\label{tab:mca_biored_novel}
\end{center}
\end{table}

%
%
\begin{table}[t]
\begin{center}
\begin{adjustbox}{max width=.48\textwidth}
\begin{tabular}{ l | c | c | c | c | c |} \cline{2-6}
                                        & \rb{90}{h(Up regulator)} & \rb{90}{h(Down regulator)} & \rb{90}{h(Agonist)} & \rb{90}{h(Antagonist)} & \rb{90}{h(Substrate)} \\ \hline
    \multicolumn{1}{|l|}{Up regulator}  & \cc{green!25}2 & \cc{red!25}0 & \cc{yellow!25}1 & \cc{yellow!25}1 & \cc{yellow!25}1 \\ \hline
    \multicolumn{1}{|l|}{Down regulator}& \cc{red!25}0  & \cc{green!25}2 & \cc{yellow!25}1 & \cc{yellow!25}1 & \cc{yellow!25}1 \\ \hline
    \multicolumn{1}{|l|}{Agonist}       & \cc{yellow!25}1  & \cc{yellow!25}1 & \cc{green!25}2 & \cc{red!25}0 & \cc{yellow!25}1 \\ \hline
    \multicolumn{1}{|l|}{Antagonist}    & \cc{yellow!25}1  & \cc{yellow!25}1 & \cc{red!25}0 & \cc{green!25}2 & \cc{yellow!25}1 \\ \hline
    \multicolumn{1}{|l|}{Substrate}     & \cc{yellow!25}1  & \cc{yellow!25}1 & \cc{yellow!25}1 & \cc{yellow!25}1 & \cc{green!25}2 \\ \hline
\end{tabular}\end{adjustbox} 
\caption{Meta-class analysis for ChemProt. ``Up regulator'' is mutually exclusive to ``down regulator'' and ``agonist'' is mutually exclusive to ``antagonist.''}
\label{tab:mca_chemprot}
\end{center}
\end{table}

%
%
\begin{table}[t]
\begin{center}
\begin{adjustbox}{max width=.48\textwidth}
\begin{tabular}{ l | c | c | c | c |} \cline{2-5}
                                        & \rb{90}{h(Advise)} & \rb{90}{h(Effect)} & \rb{90}{h(Interact)} & \rb{90}{h(Mechanism) } \\ \hline
    \multicolumn{1}{|l|}{Advise}  & \cc{green!25}2 & \cc{yellow!25}1 & \cc{yellow!25}1 & \cc{yellow!25}1  \\ \hline
    \multicolumn{1}{|l|}{Effect}& \cc{yellow!25}1  & \cc{green!25}2 & \cc{yellow!25}1 & \cc{yellow!25}1 \\ \hline
    \multicolumn{1}{|l|}{Interact}       & \cc{yellow!25}1  & \cc{yellow!25}1 & \cc{green!25}2 & \cc{yellow!25}1  \\ \hline
    \multicolumn{1}{|l|}{Mechanism}    & \cc{yellow!25}1  & \cc{yellow!25}1 & \cc{yellow!25}1 & \cc{green!25}2  \\ \hline
\end{tabular}\end{adjustbox} 
\caption{Meta-class analysis for DDI13. No classes in DDI13 are mutually exclusive based on class definitions.}
\label{tab:mca_ddi13}
\end{center}
\end{table}

%
%

%

\begin{table*}[th]
\begin{center}
\begin{adjustbox}{max width=1.0\textwidth}
\begin{tabular}{ l | c | c | c | c | c | c | c | c | c | c | c | c | c | c | c | c | c | c | c | c | c | c | c | c | c | c | c | c | c | c | c | c | c | c | c | c | c | c | c | c |} \cline{2-41}
    & \rb{90}{h(no relation)} & \rb{90}{h(org:alternate names)} & \rb{90}{h(org:city of branch)} & \rb{90}{h(org:country of branch)} & \rb{90}{h(org:dissolved)} & \rb{90}{h(org:founded)} & \rb{90}{h(org:founded by)} & \rb{90}{h(org:member of)} & \rb{90}{h(org:members)} & \rb{90}{h(org:number of employees/members )} & \rb{90}{h(org:political/religious affiliation)} & \rb{90}{h(org:shareholders)} & \rb{90}{h(org:state or province of branch)} & \rb{90}{h(org:top members/employees)} & \rb{90}{h(org:website)} & \rb{90}{h(per:age)} & \rb{90}{h(per:cause of death)} & \rb{90}{h(per:charges)} & \rb{90}{h(per:children)} & \rb{90}{h(per:cities of residence)} & \rb{90}{h(per:city of birth)} & \rb{90}{h(per:city of death)} & \rb{90}{h(per:countries of residence)} & \rb{90}{h(per:country of birth)} & \rb{90}{h(per:country of death)} & \rb{90}{h(per:date of birth)} & \rb{90}{h(per:date of death)} & \rb{90}{h(per:employee of)} & \rb{90}{h(per:identity)} & \rb{90}{h(per:origin)} & \rb{90}{h(per:other family)} & \rb{90}{h(per:parents)} & \rb{90}{h(per:religion)} & \rb{90}{h(per:schools attended)} & \rb{90}{h(per:siblings)} & \rb{90}{h(per:spouse)} & \rb{90}{h(per:state or province of birth)} & \rb{90}{h(per:state or province of death)} & \rb{90}{h(per:state or provinces of residence)} & \rb{90}{h(per:title)} \\ \hline

\multicolumn{1}{|l|}{No relation}                           &\cc{green!25}2&\cc{red!25}0&\cc{red!25}0&\cc{red!25}0&\cc{red!25}0&\cc{red!25}0&\cc{red!25}0&\cc{red!25}0&\cc{red!25}0&\cc{red!25}0&\cc{red!25}0&\cc{red!25}0&\cc{red!25}0&\cc{red!25}0&\cc{red!25}0&\cc{red!25}0&\cc{red!25}0&\cc{red!25}0&\cc{red!25}0&\cc{red!25}0&\cc{red!25}0&\cc{red!25}0&\cc{red!25}0&\cc{red!25}0&\cc{red!25}0&\cc{red!25}0&\cc{red!25}0&\cc{red!25}0&\cc{red!25}0&\cc{red!25}0&\cc{red!25}0&\cc{red!25}0&\cc{red!25}0&\cc{red!25}0&\cc{red!25}0&\cc{red!25}0&\cc{red!25}0&\cc{red!25}0&\cc{red!25}0&\cc{red!25}0\\ \hline 
\multicolumn{1}{|l|}{org:alternate names}                   &\cc{red!25}0&\cc{green!25}2&\cc{yellow!25}1&\cc{yellow!25}1&\cc{yellow!25}1&\cc{yellow!25}1&\cc{yellow!25}1&\cc{yellow!25}1&\cc{yellow!25}1&\cc{yellow!25}1&\cc{yellow!25}1&\cc{yellow!25}1&\cc{yellow!25}1&\cc{yellow!25}1&\cc{yellow!25}1&\cc{yellow!25}1&\cc{yellow!25}1&\cc{yellow!25}1&\cc{yellow!25}1&\cc{yellow!25}1&\cc{yellow!25}1&\cc{yellow!25}1&\cc{yellow!25}1&\cc{yellow!25}1&\cc{yellow!25}1&\cc{yellow!25}1&\cc{yellow!25}1&\cc{yellow!25}1&\cc{yellow!25}1&\cc{yellow!25}1&\cc{yellow!25}1&\cc{yellow!25}1&\cc{yellow!25}1&\cc{yellow!25}1&\cc{yellow!25}1&\cc{yellow!25}1&\cc{yellow!25}1&\cc{yellow!25}1&\cc{yellow!25}1&\cc{yellow!25}1\\ \hline 
\multicolumn{1}{|l|}{org:city of branch}                    &\cc{red!25}0&\cc{yellow!25}1&\cc{green!25}2&\cc{yellow!25}1&\cc{yellow!25}1&\cc{yellow!25}1&\cc{yellow!25}1&\cc{yellow!25}1&\cc{yellow!25}1&\cc{yellow!25}1&\cc{yellow!25}1&\cc{yellow!25}1&\cc{yellow!25}1&\cc{yellow!25}1&\cc{yellow!25}1&\cc{yellow!25}1&\cc{yellow!25}1&\cc{yellow!25}1&\cc{yellow!25}1&\cc{yellow!25}1&\cc{yellow!25}1&\cc{yellow!25}1&\cc{yellow!25}1&\cc{yellow!25}1&\cc{yellow!25}1&\cc{yellow!25}1&\cc{yellow!25}1&\cc{yellow!25}1&\cc{yellow!25}1&\cc{yellow!25}1&\cc{yellow!25}1&\cc{yellow!25}1&\cc{yellow!25}1&\cc{yellow!25}1&\cc{yellow!25}1&\cc{yellow!25}1&\cc{yellow!25}1&\cc{yellow!25}1&\cc{yellow!25}1&\cc{yellow!25}1\\ \hline 
\multicolumn{1}{|l|}{org:country of branch}                 &\cc{red!25}0&\cc{yellow!25}1&\cc{yellow!25}1&\cc{green!25}2&\cc{yellow!25}1&\cc{yellow!25}1&\cc{yellow!25}1&\cc{yellow!25}1&\cc{yellow!25}1&\cc{yellow!25}1&\cc{yellow!25}1&\cc{yellow!25}1&\cc{yellow!25}1&\cc{yellow!25}1&\cc{yellow!25}1&\cc{yellow!25}1&\cc{yellow!25}1&\cc{yellow!25}1&\cc{yellow!25}1&\cc{yellow!25}1&\cc{yellow!25}1&\cc{yellow!25}1&\cc{yellow!25}1&\cc{yellow!25}1&\cc{yellow!25}1&\cc{yellow!25}1&\cc{yellow!25}1&\cc{yellow!25}1&\cc{yellow!25}1&\cc{yellow!25}1&\cc{yellow!25}1&\cc{yellow!25}1&\cc{yellow!25}1&\cc{yellow!25}1&\cc{yellow!25}1&\cc{yellow!25}1&\cc{yellow!25}1&\cc{yellow!25}1&\cc{yellow!25}1&\cc{yellow!25}1\\ \hline 
\multicolumn{1}{|l|}{org:dissolved}                         &\cc{red!25}0&\cc{yellow!25}1&\cc{yellow!25}1&\cc{yellow!25}1&\cc{green!25}2&\cc{yellow!25}1&\cc{yellow!25}1&\cc{yellow!25}1&\cc{yellow!25}1&\cc{yellow!25}1&\cc{yellow!25}1&\cc{yellow!25}1&\cc{yellow!25}1&\cc{yellow!25}1&\cc{yellow!25}1&\cc{yellow!25}1&\cc{yellow!25}1&\cc{yellow!25}1&\cc{yellow!25}1&\cc{yellow!25}1&\cc{yellow!25}1&\cc{yellow!25}1&\cc{yellow!25}1&\cc{yellow!25}1&\cc{yellow!25}1&\cc{yellow!25}1&\cc{yellow!25}1&\cc{yellow!25}1&\cc{yellow!25}1&\cc{yellow!25}1&\cc{yellow!25}1&\cc{yellow!25}1&\cc{yellow!25}1&\cc{yellow!25}1&\cc{yellow!25}1&\cc{yellow!25}1&\cc{yellow!25}1&\cc{yellow!25}1&\cc{yellow!25}1&\cc{yellow!25}1\\ \hline 
\multicolumn{1}{|l|}{org:founded}                           &\cc{red!25}0&\cc{yellow!25}1&\cc{yellow!25}1&\cc{yellow!25}1&\cc{yellow!25}1&\cc{green!25}2&\cc{yellow!25}1&\cc{yellow!25}1&\cc{yellow!25}1&\cc{yellow!25}1&\cc{yellow!25}1&\cc{yellow!25}1&\cc{yellow!25}1&\cc{yellow!25}1&\cc{yellow!25}1&\cc{yellow!25}1&\cc{yellow!25}1&\cc{yellow!25}1&\cc{yellow!25}1&\cc{yellow!25}1&\cc{yellow!25}1&\cc{yellow!25}1&\cc{yellow!25}1&\cc{yellow!25}1&\cc{yellow!25}1&\cc{yellow!25}1&\cc{yellow!25}1&\cc{yellow!25}1&\cc{yellow!25}1&\cc{yellow!25}1&\cc{yellow!25}1&\cc{yellow!25}1&\cc{yellow!25}1&\cc{yellow!25}1&\cc{yellow!25}1&\cc{yellow!25}1&\cc{yellow!25}1&\cc{yellow!25}1&\cc{yellow!25}1&\cc{yellow!25}1\\ \hline 
\multicolumn{1}{|l|}{org:founded by}                        &\cc{red!25}0&\cc{yellow!25}1&\cc{yellow!25}1&\cc{yellow!25}1&\cc{yellow!25}1&\cc{yellow!25}1&\cc{green!25}2&\cc{yellow!25}1&\cc{yellow!25}1&\cc{yellow!25}1&\cc{yellow!25}1&\cc{yellow!25}1&\cc{yellow!25}1&\cc{yellow!25}1&\cc{yellow!25}1&\cc{yellow!25}1&\cc{yellow!25}1&\cc{yellow!25}1&\cc{yellow!25}1&\cc{yellow!25}1&\cc{yellow!25}1&\cc{yellow!25}1&\cc{yellow!25}1&\cc{yellow!25}1&\cc{yellow!25}1&\cc{yellow!25}1&\cc{yellow!25}1&\cc{yellow!25}1&\cc{yellow!25}1&\cc{yellow!25}1&\cc{yellow!25}1&\cc{yellow!25}1&\cc{yellow!25}1&\cc{yellow!25}1&\cc{yellow!25}1&\cc{yellow!25}1&\cc{yellow!25}1&\cc{yellow!25}1&\cc{yellow!25}1&\cc{yellow!25}1\\ \hline 
\multicolumn{1}{|l|}{org:member of}                         &\cc{red!25}0&\cc{yellow!25}1&\cc{yellow!25}1&\cc{yellow!25}1&\cc{yellow!25}1&\cc{yellow!25}1&\cc{yellow!25}1&\cc{green!25}2&\cc{red!25}0&\cc{yellow!25}1&\cc{yellow!25}1&\cc{yellow!25}1&\cc{yellow!25}1&\cc{yellow!25}1&\cc{yellow!25}1&\cc{yellow!25}1&\cc{yellow!25}1&\cc{yellow!25}1&\cc{yellow!25}1&\cc{yellow!25}1&\cc{yellow!25}1&\cc{yellow!25}1&\cc{yellow!25}1&\cc{yellow!25}1&\cc{yellow!25}1&\cc{yellow!25}1&\cc{yellow!25}1&\cc{yellow!25}1&\cc{yellow!25}1&\cc{yellow!25}1&\cc{yellow!25}1&\cc{yellow!25}1&\cc{yellow!25}1&\cc{yellow!25}1&\cc{yellow!25}1&\cc{yellow!25}1&\cc{yellow!25}1&\cc{yellow!25}1&\cc{yellow!25}1&\cc{yellow!25}1\\ \hline 
\multicolumn{1}{|l|}{org:members}                           &\cc{red!25}0&\cc{yellow!25}1&\cc{yellow!25}1&\cc{yellow!25}1&\cc{yellow!25}1&\cc{yellow!25}1&\cc{yellow!25}1&\cc{red!25}0&\cc{green!25}2&\cc{yellow!25}1&\cc{yellow!25}1&\cc{yellow!25}1&\cc{yellow!25}1&\cc{yellow!25}1&\cc{yellow!25}1&\cc{yellow!25}1&\cc{yellow!25}1&\cc{yellow!25}1&\cc{yellow!25}1&\cc{yellow!25}1&\cc{yellow!25}1&\cc{yellow!25}1&\cc{yellow!25}1&\cc{yellow!25}1&\cc{yellow!25}1&\cc{yellow!25}1&\cc{yellow!25}1&\cc{yellow!25}1&\cc{yellow!25}1&\cc{yellow!25}1&\cc{yellow!25}1&\cc{yellow!25}1&\cc{yellow!25}1&\cc{yellow!25}1&\cc{yellow!25}1&\cc{yellow!25}1&\cc{yellow!25}1&\cc{yellow!25}1&\cc{yellow!25}1&\cc{yellow!25}1\\ \hline 
\multicolumn{1}{|l|}{org:number of employees/members}       &\cc{red!25}0&\cc{yellow!25}1&\cc{yellow!25}1&\cc{yellow!25}1&\cc{yellow!25}1&\cc{yellow!25}1&\cc{yellow!25}1&\cc{yellow!25}1&\cc{yellow!25}1&\cc{green!25}2&\cc{yellow!25}1&\cc{yellow!25}1&\cc{yellow!25}1&\cc{yellow!25}1&\cc{yellow!25}1&\cc{yellow!25}1&\cc{yellow!25}1&\cc{yellow!25}1&\cc{yellow!25}1&\cc{yellow!25}1&\cc{yellow!25}1&\cc{yellow!25}1&\cc{yellow!25}1&\cc{yellow!25}1&\cc{yellow!25}1&\cc{yellow!25}1&\cc{yellow!25}1&\cc{yellow!25}1&\cc{yellow!25}1&\cc{yellow!25}1&\cc{yellow!25}1&\cc{yellow!25}1&\cc{yellow!25}1&\cc{yellow!25}1&\cc{yellow!25}1&\cc{yellow!25}1&\cc{yellow!25}1&\cc{yellow!25}1&\cc{yellow!25}1&\cc{yellow!25}1\\ \hline 
\multicolumn{1}{|l|}{org:political/religious affiliation}   &\cc{red!25}0&\cc{yellow!25}1&\cc{yellow!25}1&\cc{yellow!25}1&\cc{yellow!25}1&\cc{yellow!25}1&\cc{yellow!25}1&\cc{yellow!25}1&\cc{yellow!25}1&\cc{yellow!25}1&\cc{green!25}2&\cc{yellow!25}1&\cc{yellow!25}1&\cc{yellow!25}1&\cc{yellow!25}1&\cc{yellow!25}1&\cc{yellow!25}1&\cc{yellow!25}1&\cc{yellow!25}1&\cc{yellow!25}1&\cc{yellow!25}1&\cc{yellow!25}1&\cc{yellow!25}1&\cc{yellow!25}1&\cc{yellow!25}1&\cc{yellow!25}1&\cc{yellow!25}1&\cc{yellow!25}1&\cc{yellow!25}1&\cc{yellow!25}1&\cc{yellow!25}1&\cc{yellow!25}1&\cc{yellow!25}1&\cc{yellow!25}1&\cc{yellow!25}1&\cc{yellow!25}1&\cc{yellow!25}1&\cc{yellow!25}1&\cc{yellow!25}1&\cc{yellow!25}1\\ \hline 
\multicolumn{1}{|l|}{org:shareholders}                      &\cc{red!25}0&\cc{yellow!25}1&\cc{yellow!25}1&\cc{yellow!25}1&\cc{yellow!25}1&\cc{yellow!25}1&\cc{yellow!25}1&\cc{yellow!25}1&\cc{yellow!25}1&\cc{yellow!25}1&\cc{yellow!25}1&\cc{green!25}2&\cc{yellow!25}1&\cc{yellow!25}1&\cc{yellow!25}1&\cc{yellow!25}1&\cc{yellow!25}1&\cc{yellow!25}1&\cc{yellow!25}1&\cc{yellow!25}1&\cc{yellow!25}1&\cc{yellow!25}1&\cc{yellow!25}1&\cc{yellow!25}1&\cc{yellow!25}1&\cc{yellow!25}1&\cc{yellow!25}1&\cc{yellow!25}1&\cc{yellow!25}1&\cc{yellow!25}1&\cc{yellow!25}1&\cc{yellow!25}1&\cc{yellow!25}1&\cc{yellow!25}1&\cc{yellow!25}1&\cc{yellow!25}1&\cc{yellow!25}1&\cc{yellow!25}1&\cc{yellow!25}1&\cc{yellow!25}1\\ \hline 
\multicolumn{1}{|l|}{org:state or province of branch}       &\cc{red!25}0&\cc{yellow!25}1&\cc{yellow!25}1&\cc{yellow!25}1&\cc{yellow!25}1&\cc{yellow!25}1&\cc{yellow!25}1&\cc{yellow!25}1&\cc{yellow!25}1&\cc{yellow!25}1&\cc{yellow!25}1&\cc{yellow!25}1&\cc{green!25}2&\cc{yellow!25}1&\cc{yellow!25}1&\cc{yellow!25}1&\cc{yellow!25}1&\cc{yellow!25}1&\cc{yellow!25}1&\cc{yellow!25}1&\cc{yellow!25}1&\cc{yellow!25}1&\cc{yellow!25}1&\cc{yellow!25}1&\cc{yellow!25}1&\cc{yellow!25}1&\cc{yellow!25}1&\cc{yellow!25}1&\cc{yellow!25}1&\cc{yellow!25}1&\cc{yellow!25}1&\cc{yellow!25}1&\cc{yellow!25}1&\cc{yellow!25}1&\cc{yellow!25}1&\cc{yellow!25}1&\cc{yellow!25}1&\cc{yellow!25}1&\cc{yellow!25}1&\cc{yellow!25}1\\ \hline 
\multicolumn{1}{|l|}{org:top members/employees}             &\cc{red!25}0&\cc{yellow!25}1&\cc{yellow!25}1&\cc{yellow!25}1&\cc{yellow!25}1&\cc{yellow!25}1&\cc{yellow!25}1&\cc{yellow!25}1&\cc{yellow!25}1&\cc{yellow!25}1&\cc{yellow!25}1&\cc{yellow!25}1&\cc{yellow!25}1&\cc{green!25}2&\cc{yellow!25}1&\cc{yellow!25}1&\cc{yellow!25}1&\cc{yellow!25}1&\cc{yellow!25}1&\cc{yellow!25}1&\cc{yellow!25}1&\cc{yellow!25}1&\cc{yellow!25}1&\cc{yellow!25}1&\cc{yellow!25}1&\cc{yellow!25}1&\cc{yellow!25}1&\cc{yellow!25}1&\cc{yellow!25}1&\cc{yellow!25}1&\cc{yellow!25}1&\cc{yellow!25}1&\cc{yellow!25}1&\cc{yellow!25}1&\cc{yellow!25}1&\cc{yellow!25}1&\cc{yellow!25}1&\cc{yellow!25}1&\cc{yellow!25}1&\cc{yellow!25}1\\ \hline 
\multicolumn{1}{|l|}{org:website}                           &\cc{red!25}0&\cc{yellow!25}1&\cc{yellow!25}1&\cc{yellow!25}1&\cc{yellow!25}1&\cc{yellow!25}1&\cc{yellow!25}1&\cc{yellow!25}1&\cc{yellow!25}1&\cc{yellow!25}1&\cc{yellow!25}1&\cc{yellow!25}1&\cc{yellow!25}1&\cc{yellow!25}1&\cc{green!25}2&\cc{yellow!25}1&\cc{yellow!25}1&\cc{yellow!25}1&\cc{yellow!25}1&\cc{yellow!25}1&\cc{yellow!25}1&\cc{yellow!25}1&\cc{yellow!25}1&\cc{yellow!25}1&\cc{yellow!25}1&\cc{yellow!25}1&\cc{yellow!25}1&\cc{yellow!25}1&\cc{yellow!25}1&\cc{yellow!25}1&\cc{yellow!25}1&\cc{yellow!25}1&\cc{yellow!25}1&\cc{yellow!25}1&\cc{yellow!25}1&\cc{yellow!25}1&\cc{yellow!25}1&\cc{yellow!25}1&\cc{yellow!25}1&\cc{yellow!25}1\\ \hline 
\multicolumn{1}{|l|}{per:age}                               &\cc{red!25}0&\cc{yellow!25}1&\cc{yellow!25}1&\cc{yellow!25}1&\cc{yellow!25}1&\cc{yellow!25}1&\cc{yellow!25}1&\cc{yellow!25}1&\cc{yellow!25}1&\cc{yellow!25}1&\cc{yellow!25}1&\cc{yellow!25}1&\cc{yellow!25}1&\cc{yellow!25}1&\cc{yellow!25}1&\cc{green!25}2&\cc{yellow!25}1&\cc{yellow!25}1&\cc{yellow!25}1&\cc{yellow!25}1&\cc{yellow!25}1&\cc{yellow!25}1&\cc{yellow!25}1&\cc{yellow!25}1&\cc{yellow!25}1&\cc{yellow!25}1&\cc{yellow!25}1&\cc{yellow!25}1&\cc{yellow!25}1&\cc{yellow!25}1&\cc{yellow!25}1&\cc{yellow!25}1&\cc{yellow!25}1&\cc{yellow!25}1&\cc{yellow!25}1&\cc{yellow!25}1&\cc{yellow!25}1&\cc{yellow!25}1&\cc{yellow!25}1&\cc{yellow!25}1\\ \hline 
\multicolumn{1}{|l|}{per:cause of death}                    &\cc{red!25}0&\cc{yellow!25}1&\cc{yellow!25}1&\cc{yellow!25}1&\cc{yellow!25}1&\cc{yellow!25}1&\cc{yellow!25}1&\cc{yellow!25}1&\cc{yellow!25}1&\cc{yellow!25}1&\cc{yellow!25}1&\cc{yellow!25}1&\cc{yellow!25}1&\cc{yellow!25}1&\cc{yellow!25}1&\cc{yellow!25}1&\cc{green!25}2&\cc{yellow!25}1&\cc{yellow!25}1&\cc{yellow!25}1&\cc{yellow!25}1&\cc{yellow!25}1&\cc{yellow!25}1&\cc{yellow!25}1&\cc{yellow!25}1&\cc{yellow!25}1&\cc{yellow!25}1&\cc{yellow!25}1&\cc{yellow!25}1&\cc{yellow!25}1&\cc{yellow!25}1&\cc{yellow!25}1&\cc{yellow!25}1&\cc{yellow!25}1&\cc{yellow!25}1&\cc{yellow!25}1&\cc{yellow!25}1&\cc{yellow!25}1&\cc{yellow!25}1&\cc{yellow!25}1\\ \hline 
\multicolumn{1}{|l|}{per:charges}                           &\cc{red!25}0&\cc{yellow!25}1&\cc{yellow!25}1&\cc{yellow!25}1&\cc{yellow!25}1&\cc{yellow!25}1&\cc{yellow!25}1&\cc{yellow!25}1&\cc{yellow!25}1&\cc{yellow!25}1&\cc{yellow!25}1&\cc{yellow!25}1&\cc{yellow!25}1&\cc{yellow!25}1&\cc{yellow!25}1&\cc{yellow!25}1&\cc{yellow!25}1&\cc{green!25}2&\cc{yellow!25}1&\cc{yellow!25}1&\cc{yellow!25}1&\cc{yellow!25}1&\cc{yellow!25}1&\cc{yellow!25}1&\cc{yellow!25}1&\cc{yellow!25}1&\cc{yellow!25}1&\cc{yellow!25}1&\cc{yellow!25}1&\cc{yellow!25}1&\cc{yellow!25}1&\cc{yellow!25}1&\cc{yellow!25}1&\cc{yellow!25}1&\cc{yellow!25}1&\cc{yellow!25}1&\cc{yellow!25}1&\cc{yellow!25}1&\cc{yellow!25}1&\cc{yellow!25}1\\ \hline 
\multicolumn{1}{|l|}{per:children}                          &\cc{red!25}0&\cc{yellow!25}1&\cc{yellow!25}1&\cc{yellow!25}1&\cc{yellow!25}1&\cc{yellow!25}1&\cc{yellow!25}1&\cc{yellow!25}1&\cc{yellow!25}1&\cc{yellow!25}1&\cc{yellow!25}1&\cc{yellow!25}1&\cc{yellow!25}1&\cc{yellow!25}1&\cc{yellow!25}1&\cc{yellow!25}1&\cc{yellow!25}1&\cc{yellow!25}1&\cc{green!25}2&\cc{yellow!25}1&\cc{yellow!25}1&\cc{yellow!25}1&\cc{yellow!25}1&\cc{yellow!25}1&\cc{yellow!25}1&\cc{yellow!25}1&\cc{yellow!25}1&\cc{yellow!25}1&\cc{red!25}0&\cc{yellow!25}1&\cc{red!25}0&\cc{red!25}0&\cc{yellow!25}1&\cc{yellow!25}1&\cc{red!25}0&\cc{red!25}0&\cc{yellow!25}1&\cc{yellow!25}1&\cc{yellow!25}1&\cc{yellow!25}1\\ \hline 
\multicolumn{1}{|l|}{per:cities of residence}               &\cc{red!25}0&\cc{yellow!25}1&\cc{yellow!25}1&\cc{yellow!25}1&\cc{yellow!25}1&\cc{yellow!25}1&\cc{yellow!25}1&\cc{yellow!25}1&\cc{yellow!25}1&\cc{yellow!25}1&\cc{yellow!25}1&\cc{yellow!25}1&\cc{yellow!25}1&\cc{yellow!25}1&\cc{yellow!25}1&\cc{yellow!25}1&\cc{yellow!25}1&\cc{yellow!25}1&\cc{yellow!25}1&\cc{green!25}2&\cc{yellow!25}1&\cc{yellow!25}1&\cc{yellow!25}1&\cc{yellow!25}1&\cc{yellow!25}1&\cc{yellow!25}1&\cc{yellow!25}1&\cc{yellow!25}1&\cc{yellow!25}1&\cc{yellow!25}1&\cc{yellow!25}1&\cc{yellow!25}1&\cc{yellow!25}1&\cc{yellow!25}1&\cc{yellow!25}1&\cc{yellow!25}1&\cc{yellow!25}1&\cc{yellow!25}1&\cc{yellow!25}1&\cc{yellow!25}1\\ \hline 
\multicolumn{1}{|l|}{per:city of birth}                     &\cc{red!25}0&\cc{yellow!25}1&\cc{yellow!25}1&\cc{yellow!25}1&\cc{yellow!25}1&\cc{yellow!25}1&\cc{yellow!25}1&\cc{yellow!25}1&\cc{yellow!25}1&\cc{yellow!25}1&\cc{yellow!25}1&\cc{yellow!25}1&\cc{yellow!25}1&\cc{yellow!25}1&\cc{yellow!25}1&\cc{yellow!25}1&\cc{yellow!25}1&\cc{yellow!25}1&\cc{yellow!25}1&\cc{yellow!25}1&\cc{green!25}2&\cc{yellow!25}1&\cc{yellow!25}1&\cc{yellow!25}1&\cc{yellow!25}1&\cc{yellow!25}1&\cc{yellow!25}1&\cc{yellow!25}1&\cc{yellow!25}1&\cc{yellow!25}1&\cc{yellow!25}1&\cc{yellow!25}1&\cc{yellow!25}1&\cc{yellow!25}1&\cc{yellow!25}1&\cc{yellow!25}1&\cc{yellow!25}1&\cc{yellow!25}1&\cc{yellow!25}1&\cc{yellow!25}1\\ \hline 
\multicolumn{1}{|l|}{per:city of death}                     &\cc{red!25}0&\cc{yellow!25}1&\cc{yellow!25}1&\cc{yellow!25}1&\cc{yellow!25}1&\cc{yellow!25}1&\cc{yellow!25}1&\cc{yellow!25}1&\cc{yellow!25}1&\cc{yellow!25}1&\cc{yellow!25}1&\cc{yellow!25}1&\cc{yellow!25}1&\cc{yellow!25}1&\cc{yellow!25}1&\cc{yellow!25}1&\cc{yellow!25}1&\cc{yellow!25}1&\cc{yellow!25}1&\cc{yellow!25}1&\cc{yellow!25}1&\cc{green!25}2&\cc{yellow!25}1&\cc{yellow!25}1&\cc{yellow!25}1&\cc{yellow!25}1&\cc{yellow!25}1&\cc{yellow!25}1&\cc{yellow!25}1&\cc{yellow!25}1&\cc{yellow!25}1&\cc{yellow!25}1&\cc{yellow!25}1&\cc{yellow!25}1&\cc{yellow!25}1&\cc{yellow!25}1&\cc{yellow!25}1&\cc{yellow!25}1&\cc{yellow!25}1&\cc{yellow!25}1\\ \hline 
\multicolumn{1}{|l|}{per:countries of residence}            &\cc{red!25}0&\cc{yellow!25}1&\cc{yellow!25}1&\cc{yellow!25}1&\cc{yellow!25}1&\cc{yellow!25}1&\cc{yellow!25}1&\cc{yellow!25}1&\cc{yellow!25}1&\cc{yellow!25}1&\cc{yellow!25}1&\cc{yellow!25}1&\cc{yellow!25}1&\cc{yellow!25}1&\cc{yellow!25}1&\cc{yellow!25}1&\cc{yellow!25}1&\cc{yellow!25}1&\cc{yellow!25}1&\cc{yellow!25}1&\cc{yellow!25}1&\cc{yellow!25}1&\cc{green!25}2&\cc{yellow!25}1&\cc{yellow!25}1&\cc{yellow!25}1&\cc{yellow!25}1&\cc{yellow!25}1&\cc{yellow!25}1&\cc{yellow!25}1&\cc{yellow!25}1&\cc{yellow!25}1&\cc{yellow!25}1&\cc{yellow!25}1&\cc{yellow!25}1&\cc{yellow!25}1&\cc{yellow!25}1&\cc{yellow!25}1&\cc{yellow!25}1&\cc{yellow!25}1\\ \hline 
\multicolumn{1}{|l|}{per:country of birth}                  &\cc{red!25}0&\cc{yellow!25}1&\cc{yellow!25}1&\cc{yellow!25}1&\cc{yellow!25}1&\cc{yellow!25}1&\cc{yellow!25}1&\cc{yellow!25}1&\cc{yellow!25}1&\cc{yellow!25}1&\cc{yellow!25}1&\cc{yellow!25}1&\cc{yellow!25}1&\cc{yellow!25}1&\cc{yellow!25}1&\cc{yellow!25}1&\cc{yellow!25}1&\cc{yellow!25}1&\cc{yellow!25}1&\cc{yellow!25}1&\cc{yellow!25}1&\cc{yellow!25}1&\cc{yellow!25}1&\cc{green!25}2&\cc{yellow!25}1&\cc{yellow!25}1&\cc{yellow!25}1&\cc{yellow!25}1&\cc{yellow!25}1&\cc{yellow!25}1&\cc{yellow!25}1&\cc{yellow!25}1&\cc{yellow!25}1&\cc{yellow!25}1&\cc{yellow!25}1&\cc{yellow!25}1&\cc{yellow!25}1&\cc{yellow!25}1&\cc{yellow!25}1&\cc{yellow!25}1\\ \hline 
\multicolumn{1}{|l|}{per:country of death}                  &\cc{red!25}0&\cc{yellow!25}1&\cc{yellow!25}1&\cc{yellow!25}1&\cc{yellow!25}1&\cc{yellow!25}1&\cc{yellow!25}1&\cc{yellow!25}1&\cc{yellow!25}1&\cc{yellow!25}1&\cc{yellow!25}1&\cc{yellow!25}1&\cc{yellow!25}1&\cc{yellow!25}1&\cc{yellow!25}1&\cc{yellow!25}1&\cc{yellow!25}1&\cc{yellow!25}1&\cc{yellow!25}1&\cc{yellow!25}1&\cc{yellow!25}1&\cc{yellow!25}1&\cc{yellow!25}1&\cc{yellow!25}1&\cc{green!25}2&\cc{yellow!25}1&\cc{yellow!25}1&\cc{yellow!25}1&\cc{yellow!25}1&\cc{yellow!25}1&\cc{yellow!25}1&\cc{yellow!25}1&\cc{yellow!25}1&\cc{yellow!25}1&\cc{yellow!25}1&\cc{yellow!25}1&\cc{yellow!25}1&\cc{yellow!25}1&\cc{yellow!25}1&\cc{yellow!25}1\\ \hline 
\multicolumn{1}{|l|}{per:date of birth}                     &\cc{red!25}0&\cc{yellow!25}1&\cc{yellow!25}1&\cc{yellow!25}1&\cc{yellow!25}1&\cc{yellow!25}1&\cc{yellow!25}1&\cc{yellow!25}1&\cc{yellow!25}1&\cc{yellow!25}1&\cc{yellow!25}1&\cc{yellow!25}1&\cc{yellow!25}1&\cc{yellow!25}1&\cc{yellow!25}1&\cc{yellow!25}1&\cc{yellow!25}1&\cc{yellow!25}1&\cc{yellow!25}1&\cc{yellow!25}1&\cc{yellow!25}1&\cc{yellow!25}1&\cc{yellow!25}1&\cc{yellow!25}1&\cc{yellow!25}1&\cc{green!25}2&\cc{yellow!25}1&\cc{yellow!25}1&\cc{yellow!25}1&\cc{yellow!25}1&\cc{yellow!25}1&\cc{yellow!25}1&\cc{yellow!25}1&\cc{yellow!25}1&\cc{yellow!25}1&\cc{yellow!25}1&\cc{yellow!25}1&\cc{yellow!25}1&\cc{yellow!25}1&\cc{yellow!25}1\\ \hline 
\multicolumn{1}{|l|}{per:date of death}                     &\cc{red!25}0&\cc{yellow!25}1&\cc{yellow!25}1&\cc{yellow!25}1&\cc{yellow!25}1&\cc{yellow!25}1&\cc{yellow!25}1&\cc{yellow!25}1&\cc{yellow!25}1&\cc{yellow!25}1&\cc{yellow!25}1&\cc{yellow!25}1&\cc{yellow!25}1&\cc{yellow!25}1&\cc{yellow!25}1&\cc{yellow!25}1&\cc{yellow!25}1&\cc{yellow!25}1&\cc{yellow!25}1&\cc{yellow!25}1&\cc{yellow!25}1&\cc{yellow!25}1&\cc{yellow!25}1&\cc{yellow!25}1&\cc{yellow!25}1&\cc{yellow!25}1&\cc{green!25}2&\cc{yellow!25}1&\cc{yellow!25}1&\cc{yellow!25}1&\cc{yellow!25}1&\cc{yellow!25}1&\cc{yellow!25}1&\cc{yellow!25}1&\cc{yellow!25}1&\cc{yellow!25}1&\cc{yellow!25}1&\cc{yellow!25}1&\cc{yellow!25}1&\cc{yellow!25}1\\ \hline 
\multicolumn{1}{|l|}{per:employee of}                       &\cc{red!25}0&\cc{yellow!25}1&\cc{yellow!25}1&\cc{yellow!25}1&\cc{yellow!25}1&\cc{yellow!25}1&\cc{yellow!25}1&\cc{yellow!25}1&\cc{yellow!25}1&\cc{yellow!25}1&\cc{yellow!25}1&\cc{yellow!25}1&\cc{yellow!25}1&\cc{yellow!25}1&\cc{yellow!25}1&\cc{yellow!25}1&\cc{yellow!25}1&\cc{yellow!25}1&\cc{yellow!25}1&\cc{yellow!25}1&\cc{yellow!25}1&\cc{yellow!25}1&\cc{yellow!25}1&\cc{yellow!25}1&\cc{yellow!25}1&\cc{yellow!25}1&\cc{yellow!25}1&\cc{green!25}2&\cc{yellow!25}1&\cc{yellow!25}1&\cc{yellow!25}1&\cc{yellow!25}1&\cc{yellow!25}1&\cc{yellow!25}1&\cc{yellow!25}1&\cc{yellow!25}1&\cc{yellow!25}1&\cc{yellow!25}1&\cc{yellow!25}1&\cc{yellow!25}1\\ \hline 
\multicolumn{1}{|l|}{per:identity}                          &\cc{red!25}0&\cc{yellow!25}1&\cc{yellow!25}1&\cc{yellow!25}1&\cc{yellow!25}1&\cc{yellow!25}1&\cc{yellow!25}1&\cc{yellow!25}1&\cc{yellow!25}1&\cc{yellow!25}1&\cc{yellow!25}1&\cc{yellow!25}1&\cc{yellow!25}1&\cc{yellow!25}1&\cc{yellow!25}1&\cc{yellow!25}1&\cc{yellow!25}1&\cc{yellow!25}1&\cc{red!25}0&\cc{yellow!25}1&\cc{yellow!25}1&\cc{yellow!25}1&\cc{yellow!25}1&\cc{yellow!25}1&\cc{yellow!25}1&\cc{yellow!25}1&\cc{yellow!25}1&\cc{yellow!25}1&\cc{green!25}2&\cc{yellow!25}1&\cc{red!25}0&\cc{red!25}0&\cc{yellow!25}1&\cc{yellow!25}1&\cc{red!25}0&\cc{red!25}0&\cc{yellow!25}1&\cc{yellow!25}1&\cc{yellow!25}1&\cc{yellow!25}1\\ \hline 
\multicolumn{1}{|l|}{per:origin}                            &\cc{red!25}0&\cc{yellow!25}1&\cc{yellow!25}1&\cc{yellow!25}1&\cc{yellow!25}1&\cc{yellow!25}1&\cc{yellow!25}1&\cc{yellow!25}1&\cc{yellow!25}1&\cc{yellow!25}1&\cc{yellow!25}1&\cc{yellow!25}1&\cc{yellow!25}1&\cc{yellow!25}1&\cc{yellow!25}1&\cc{yellow!25}1&\cc{yellow!25}1&\cc{yellow!25}1&\cc{yellow!25}1&\cc{yellow!25}1&\cc{yellow!25}1&\cc{yellow!25}1&\cc{yellow!25}1&\cc{yellow!25}1&\cc{yellow!25}1&\cc{yellow!25}1&\cc{yellow!25}1&\cc{yellow!25}1&\cc{yellow!25}1&\cc{green!25}2&\cc{yellow!25}1&\cc{yellow!25}1&\cc{yellow!25}1&\cc{yellow!25}1&\cc{yellow!25}1&\cc{yellow!25}1&\cc{yellow!25}1&\cc{yellow!25}1&\cc{yellow!25}1&\cc{yellow!25}1\\ \hline 
\multicolumn{1}{|l|}{per:other family}                      &\cc{red!25}0&\cc{yellow!25}1&\cc{yellow!25}1&\cc{yellow!25}1&\cc{yellow!25}1&\cc{yellow!25}1&\cc{yellow!25}1&\cc{yellow!25}1&\cc{yellow!25}1&\cc{yellow!25}1&\cc{yellow!25}1&\cc{yellow!25}1&\cc{yellow!25}1&\cc{yellow!25}1&\cc{yellow!25}1&\cc{yellow!25}1&\cc{yellow!25}1&\cc{yellow!25}1&\cc{red!25}0&\cc{yellow!25}1&\cc{yellow!25}1&\cc{yellow!25}1&\cc{yellow!25}1&\cc{yellow!25}1&\cc{yellow!25}1&\cc{yellow!25}1&\cc{yellow!25}1&\cc{yellow!25}1&\cc{red!25}0&\cc{yellow!25}1&\cc{green!25}2&\cc{red!25}0&\cc{yellow!25}1&\cc{yellow!25}1&\cc{red!25}0&\cc{red!25}0&\cc{yellow!25}1&\cc{yellow!25}1&\cc{yellow!25}1&\cc{yellow!25}1\\ \hline 
\multicolumn{1}{|l|}{per:parents}                           &\cc{red!25}0&\cc{yellow!25}1&\cc{yellow!25}1&\cc{yellow!25}1&\cc{yellow!25}1&\cc{yellow!25}1&\cc{yellow!25}1&\cc{yellow!25}1&\cc{yellow!25}1&\cc{yellow!25}1&\cc{yellow!25}1&\cc{yellow!25}1&\cc{yellow!25}1&\cc{yellow!25}1&\cc{yellow!25}1&\cc{yellow!25}1&\cc{yellow!25}1&\cc{yellow!25}1&\cc{red!25}0&\cc{yellow!25}1&\cc{yellow!25}1&\cc{yellow!25}1&\cc{yellow!25}1&\cc{yellow!25}1&\cc{yellow!25}1&\cc{yellow!25}1&\cc{yellow!25}1&\cc{yellow!25}1&\cc{red!25}0&\cc{yellow!25}1&\cc{red!25}0&\cc{green!25}2&\cc{yellow!25}1&\cc{yellow!25}1&\cc{red!25}0&\cc{red!25}0&\cc{yellow!25}1&\cc{yellow!25}1&\cc{yellow!25}1&\cc{yellow!25}1\\ \hline 
\multicolumn{1}{|l|}{per:religion}                          &\cc{red!25}0&\cc{yellow!25}1&\cc{yellow!25}1&\cc{yellow!25}1&\cc{yellow!25}1&\cc{yellow!25}1&\cc{yellow!25}1&\cc{yellow!25}1&\cc{yellow!25}1&\cc{yellow!25}1&\cc{yellow!25}1&\cc{yellow!25}1&\cc{yellow!25}1&\cc{yellow!25}1&\cc{yellow!25}1&\cc{yellow!25}1&\cc{yellow!25}1&\cc{yellow!25}1&\cc{yellow!25}1&\cc{yellow!25}1&\cc{yellow!25}1&\cc{yellow!25}1&\cc{yellow!25}1&\cc{yellow!25}1&\cc{yellow!25}1&\cc{yellow!25}1&\cc{yellow!25}1&\cc{yellow!25}1&\cc{yellow!25}1&\cc{yellow!25}1&\cc{yellow!25}1&\cc{yellow!25}1&\cc{green!25}2&\cc{yellow!25}1&\cc{yellow!25}1&\cc{yellow!25}1&\cc{yellow!25}1&\cc{yellow!25}1&\cc{yellow!25}1&\cc{yellow!25}1\\ \hline 
\multicolumn{1}{|l|}{per:schools attended}                  &\cc{red!25}0&\cc{yellow!25}1&\cc{yellow!25}1&\cc{yellow!25}1&\cc{yellow!25}1&\cc{yellow!25}1&\cc{yellow!25}1&\cc{yellow!25}1&\cc{yellow!25}1&\cc{yellow!25}1&\cc{yellow!25}1&\cc{yellow!25}1&\cc{yellow!25}1&\cc{yellow!25}1&\cc{yellow!25}1&\cc{yellow!25}1&\cc{yellow!25}1&\cc{yellow!25}1&\cc{yellow!25}1&\cc{yellow!25}1&\cc{yellow!25}1&\cc{yellow!25}1&\cc{yellow!25}1&\cc{yellow!25}1&\cc{yellow!25}1&\cc{yellow!25}1&\cc{yellow!25}1&\cc{yellow!25}1&\cc{yellow!25}1&\cc{yellow!25}1&\cc{yellow!25}1&\cc{yellow!25}1&\cc{yellow!25}1&\cc{green!25}2&\cc{yellow!25}1&\cc{yellow!25}1&\cc{yellow!25}1&\cc{yellow!25}1&\cc{yellow!25}1&\cc{yellow!25}1\\ \hline 
\multicolumn{1}{|l|}{per:siblings}                          &\cc{red!25}0&\cc{yellow!25}1&\cc{yellow!25}1&\cc{yellow!25}1&\cc{yellow!25}1&\cc{yellow!25}1&\cc{yellow!25}1&\cc{yellow!25}1&\cc{yellow!25}1&\cc{yellow!25}1&\cc{yellow!25}1&\cc{yellow!25}1&\cc{yellow!25}1&\cc{yellow!25}1&\cc{yellow!25}1&\cc{yellow!25}1&\cc{yellow!25}1&\cc{yellow!25}1&\cc{red!25}0&\cc{yellow!25}1&\cc{yellow!25}1&\cc{yellow!25}1&\cc{yellow!25}1&\cc{yellow!25}1&\cc{yellow!25}1&\cc{yellow!25}1&\cc{yellow!25}1&\cc{yellow!25}1&\cc{red!25}0&\cc{yellow!25}1&\cc{red!25}0&\cc{red!25}0&\cc{yellow!25}1&\cc{yellow!25}1&\cc{green!25}2&\cc{red!25}0&\cc{yellow!25}1&\cc{yellow!25}1&\cc{yellow!25}1&\cc{yellow!25}1\\ \hline 
\multicolumn{1}{|l|}{per:spouse}                            &\cc{red!25}0&\cc{yellow!25}1&\cc{yellow!25}1&\cc{yellow!25}1&\cc{yellow!25}1&\cc{yellow!25}1&\cc{yellow!25}1&\cc{yellow!25}1&\cc{yellow!25}1&\cc{yellow!25}1&\cc{yellow!25}1&\cc{yellow!25}1&\cc{yellow!25}1&\cc{yellow!25}1&\cc{yellow!25}1&\cc{yellow!25}1&\cc{yellow!25}1&\cc{yellow!25}1&\cc{red!25}0&\cc{yellow!25}1&\cc{yellow!25}1&\cc{yellow!25}1&\cc{yellow!25}1&\cc{yellow!25}1&\cc{yellow!25}1&\cc{yellow!25}1&\cc{yellow!25}1&\cc{yellow!25}1&\cc{red!25}0&\cc{yellow!25}1&\cc{red!25}0&\cc{red!25}0&\cc{yellow!25}1&\cc{yellow!25}1&\cc{red!25}0&\cc{green!25}2&\cc{yellow!25}1&\cc{yellow!25}1&\cc{yellow!25}1&\cc{yellow!25}1\\ \hline 
\multicolumn{1}{|l|}{per:state or province of birth}        &\cc{red!25}0&\cc{yellow!25}1&\cc{yellow!25}1&\cc{yellow!25}1&\cc{yellow!25}1&\cc{yellow!25}1&\cc{yellow!25}1&\cc{yellow!25}1&\cc{yellow!25}1&\cc{yellow!25}1&\cc{yellow!25}1&\cc{yellow!25}1&\cc{yellow!25}1&\cc{yellow!25}1&\cc{yellow!25}1&\cc{yellow!25}1&\cc{yellow!25}1&\cc{yellow!25}1&\cc{yellow!25}1&\cc{yellow!25}1&\cc{yellow!25}1&\cc{yellow!25}1&\cc{yellow!25}1&\cc{yellow!25}1&\cc{yellow!25}1&\cc{yellow!25}1&\cc{yellow!25}1&\cc{yellow!25}1&\cc{yellow!25}1&\cc{yellow!25}1&\cc{yellow!25}1&\cc{yellow!25}1&\cc{yellow!25}1&\cc{yellow!25}1&\cc{yellow!25}1&\cc{yellow!25}1&\cc{green!25}2&\cc{yellow!25}1&\cc{yellow!25}1&\cc{yellow!25}1\\ \hline 
\multicolumn{1}{|l|}{per:state or province of death}        &\cc{red!25}0&\cc{yellow!25}1&\cc{yellow!25}1&\cc{yellow!25}1&\cc{yellow!25}1&\cc{yellow!25}1&\cc{yellow!25}1&\cc{yellow!25}1&\cc{yellow!25}1&\cc{yellow!25}1&\cc{yellow!25}1&\cc{yellow!25}1&\cc{yellow!25}1&\cc{yellow!25}1&\cc{yellow!25}1&\cc{yellow!25}1&\cc{yellow!25}1&\cc{yellow!25}1&\cc{yellow!25}1&\cc{yellow!25}1&\cc{yellow!25}1&\cc{yellow!25}1&\cc{yellow!25}1&\cc{yellow!25}1&\cc{yellow!25}1&\cc{yellow!25}1&\cc{yellow!25}1&\cc{yellow!25}1&\cc{yellow!25}1&\cc{yellow!25}1&\cc{yellow!25}1&\cc{yellow!25}1&\cc{yellow!25}1&\cc{yellow!25}1&\cc{yellow!25}1&\cc{yellow!25}1&\cc{yellow!25}1&\cc{green!25}2&\cc{yellow!25}1&\cc{yellow!25}1\\ \hline 
\multicolumn{1}{|l|}{per:state or provinces of residence}   &\cc{red!25}0&\cc{yellow!25}1&\cc{yellow!25}1&\cc{yellow!25}1&\cc{yellow!25}1&\cc{yellow!25}1&\cc{yellow!25}1&\cc{yellow!25}1&\cc{yellow!25}1&\cc{yellow!25}1&\cc{yellow!25}1&\cc{yellow!25}1&\cc{yellow!25}1&\cc{yellow!25}1&\cc{yellow!25}1&\cc{yellow!25}1&\cc{yellow!25}1&\cc{yellow!25}1&\cc{yellow!25}1&\cc{yellow!25}1&\cc{yellow!25}1&\cc{yellow!25}1&\cc{yellow!25}1&\cc{yellow!25}1&\cc{yellow!25}1&\cc{yellow!25}1&\cc{yellow!25}1&\cc{yellow!25}1&\cc{yellow!25}1&\cc{yellow!25}1&\cc{yellow!25}1&\cc{yellow!25}1&\cc{yellow!25}1&\cc{yellow!25}1&\cc{yellow!25}1&\cc{yellow!25}1&\cc{yellow!25}1&\cc{yellow!25}1&\cc{green!25}2&\cc{yellow!25}1\\ \hline 
\multicolumn{1}{|l|}{per:title}                             &\cc{red!25}0&\cc{yellow!25}1&\cc{yellow!25}1&\cc{yellow!25}1&\cc{yellow!25}1&\cc{yellow!25}1&\cc{yellow!25}1&\cc{yellow!25}1&\cc{yellow!25}1&\cc{yellow!25}1&\cc{yellow!25}1&\cc{yellow!25}1&\cc{yellow!25}1&\cc{yellow!25}1&\cc{yellow!25}1&\cc{yellow!25}1&\cc{yellow!25}1&\cc{yellow!25}1&\cc{yellow!25}1&\cc{yellow!25}1&\cc{yellow!25}1&\cc{yellow!25}1&\cc{yellow!25}1&\cc{yellow!25}1&\cc{yellow!25}1&\cc{yellow!25}1&\cc{yellow!25}1&\cc{yellow!25}1&\cc{yellow!25}1&\cc{yellow!25}1&\cc{yellow!25}1&\cc{yellow!25}1&\cc{yellow!25}1&\cc{yellow!25}1&\cc{yellow!25}1&\cc{yellow!25}1&\cc{yellow!25}1&\cc{yellow!25}1&\cc{yellow!25}1&\cc{green!25}2\\ \hline 

\end{tabular}\end{adjustbox} 
\caption{Meta-class analysis for ReTACRED. Classes involving familial relations are all mutually exclusive to each other (e.g., ``per:spouse,'' ``per:parents,'' ``per:other family,'' ``per:siblings,'' ``per:identity,'' ``per:children''). Classes ``org:members'' and ``org:member of'' are mutually exclusive since each denotes an opposing directional relationship between a subject and an object.}
\label{tab:mca_retacred}
\end{center}
\end{table*}

%
%
\begin{table}[t]
\begin{center}
\begin{adjustbox}{max width=.48\textwidth}
\begin{tabular}{ l | c | c | c | c | c | c | c | c | c | c |} \cline{2-11}
                                        & \rb{90}{h(Other)} & \rb{90}{h(Component-Whole)} &  \rb{90}{h(Instrument-Agency)} & \rb{90}{h(Member-Collection)} &  \rb{90}{h(Cause-Effect)} &  \rb{90}{h(Entity-Destination)} &  \rb{90}{h(Message-Topic)} &  \rb{90}{h(Entity-Origin)} &  \rb{90}{h(Product-Producer)} &  \rb{90}{h(Content-Container)} \\ \hline
\multicolumn{1}{|l|}{Other}             &\cc{green!25}2&\cc{yellow!25}1&\cc{yellow!25}1&\cc{yellow!25}1&\cc{yellow!25}1&\cc{yellow!25}1&\cc{yellow!25}1&\cc{yellow!25}1&\cc{yellow!25}1&\cc{yellow!25}1\\ \hline
\multicolumn{1}{|l|}{Component-Whole}   &\cc{yellow!25}1&\cc{green!25}2&\cc{yellow!25}1&\cc{yellow!25}1&\cc{yellow!25}1&\cc{yellow!25}1&\cc{yellow!25}1&\cc{yellow!25}1&\cc{yellow!25}1&\cc{yellow!25}1\\ \hline
\multicolumn{1}{|l|}{Instrument-Agency} &\cc{yellow!25}1&\cc{yellow!25}1&\cc{green!25}2&\cc{yellow!25}1&\cc{yellow!25}1&\cc{yellow!25}1&\cc{yellow!25}1&\cc{yellow!25}1&\cc{yellow!25}1&\cc{yellow!25}1\\ \hline
\multicolumn{1}{|l|}{Member-Collection} &\cc{yellow!25}1&\cc{yellow!25}1&\cc{yellow!25}1&\cc{green!25}2&\cc{yellow!25}1&\cc{yellow!25}1&\cc{yellow!25}1&\cc{yellow!25}1&\cc{yellow!25}1&\cc{yellow!25}1\\ \hline
\multicolumn{1}{|l|}{Cause-Effect}      &\cc{yellow!25}1&\cc{yellow!25}1&\cc{yellow!25}1&\cc{yellow!25}1&\cc{green!25}2&\cc{yellow!25}1&\cc{yellow!25}1&\cc{yellow!25}1&\cc{yellow!25}1&\cc{yellow!25}1\\ \hline
\multicolumn{1}{|l|}{Entity-Destination}&\cc{yellow!25}1&\cc{yellow!25}1&\cc{yellow!25}1&\cc{yellow!25}1&\cc{yellow!25}1&\cc{green!25}2&\cc{yellow!25}1&\cc{yellow!25}1&\cc{yellow!25}1&\cc{yellow!25}1\\ \hline
\multicolumn{1}{|l|}{Message-Topic}     &\cc{yellow!25}1&\cc{yellow!25}1&\cc{yellow!25}1&\cc{yellow!25}1&\cc{yellow!25}1&\cc{yellow!25}1&\cc{green!25}2&\cc{yellow!25}1&\cc{yellow!25}1&\cc{yellow!25}1\\ \hline
\multicolumn{1}{|l|}{Entity-Origin}     &\cc{yellow!25}1&\cc{yellow!25}1&\cc{yellow!25}1&\cc{yellow!25}1&\cc{yellow!25}1&\cc{yellow!25}1&\cc{yellow!25}1&\cc{green!25}2&\cc{yellow!25}1&\cc{yellow!25}1\\ \hline
\multicolumn{1}{|l|}{Product-Producer}  &\cc{yellow!25}1&\cc{yellow!25}1&\cc{yellow!25}1&\cc{yellow!25}1&\cc{yellow!25}1&\cc{yellow!25}1&\cc{yellow!25}1&\cc{yellow!25}1&\cc{green!25}2&\cc{yellow!25}1\\ \hline
\multicolumn{1}{|l|}{Content-Container} &\cc{yellow!25}1&\cc{yellow!25}1&\cc{yellow!25}1&\cc{yellow!25}1&\cc{yellow!25}1&\cc{yellow!25}1&\cc{yellow!25}1&\cc{yellow!25}1&\cc{yellow!25}1&\cc{green!25}2\\ \hline
\end{tabular}\end{adjustbox}            
\caption{Meta-class analysis for SemEval-2010 Task 8. No classes in SemEval-2010 Task 8 are mutually exclusive based on class definitions.}
\label{tab:mca_semeval}
\end{center}
\end{table}

%
%
\begin{table}[t]
\begin{center}
\begin{adjustbox}{max width=.48\textwidth}
\begin{tabular}{ l | c | c |} \cline{2-3}
    &	h(Associated) & h(Not Associated) \\ \hline
    \multicolumn{1}{|l|}{Associated}          & \cc{green!25}2 & \cc{red!25}0 \\ \hline
    \multicolumn{1}{|l|}{Not Associated}      & \cc{red!25}0 & \cc{green!25}2 \\ \hline
\end{tabular}\end{adjustbox} 
\caption{Meta-class analysis for GAD. The ``Associated'' class is definitionally mutually exclusive to the ``Not Associated'' class.}
\label{tab:mca_gad}
\end{center}
\end{table}

\end{document}